\documentclass[preprint,review,12pt]{elsarticle}

\usepackage[utf8]{inputenc}
\usepackage[T1]{fontenc}

\usepackage{amssymb}
\usepackage{amsmath}
\usepackage{amsfonts}

\usepackage{bbm}

\usepackage{array}
\usepackage{color, colortbl}
\usepackage{xcolor}
\usepackage{graphicx}
\usepackage{tabularx}
\usepackage{booktabs}
\usepackage{multirow}
\usepackage{makecell}
\usepackage{caption}
\usepackage{adjustbox}

\usepackage{hyperref}
\hypersetup{
  colorlinks=false,
  linkbordercolor=white,
  urlbordercolor=white,
  pdfborder={0 0 0}
}

\usepackage[titletoc]{appendix}

\usepackage{tikz} 
\usepackage{pgfplots}
\pgfplotsset{compat=1.18}	
\usetikzlibrary{pgfplots.statistics} 
\usepgfplotslibrary{groupplots}

\usetikzlibrary{positioning, calc, matrix, decorations, decorations.pathreplacing, decorations.pathmorphing, calligraphy, arrows.meta}
\usetikzlibrary{
    shapes,
    shapes.geometric,
    shapes.symbols,
    shapes.arrows,
    shapes.multipart,
    shapes.callouts,
    shapes.misc}
\tikzset{>=Stealth[round]}
\tikzset{
    ncbar angle/.initial=90,
    ncbar/.style={
        to path=(\tikztostart)
        -- ($(\tikztostart)!#1!\pgfkeysvalueof{/tikz/ncbar angle}:(\tikztotarget)$)
        -- ($(\tikztotarget)!($(\tikztostart)!#1!\pgfkeysvalueof{/tikz/ncbar angle}:(\tikztotarget)$)!\pgfkeysvalueof{/tikz/ncbar angle}:(\tikztostart)$)
        -- (\tikztotarget)
    },
    ncbar/.default=0.5cm,
}

\tikzset{square left brace/.style={ncbar=0.2cm}}
\tikzset{square right brace/.style={ncbar=-0.2cm}}

\makeatletter
\renewcommand*\env@matrix[1][*\c@MaxMatrixCols c]{%
  \hskip -\arraycolsep
  \let\@ifnextchar\new@ifnextchar
  \array{#1}}

\makeatother
\newcolumntype{C}[1]{>{\hspace{0pt}\centering\arraybackslash}p{#1}}

\journal{Computers in Biology and Medicine}

\usepackage{xcolor}

\newcommand{\modelName}{MARIA}
\newcommand{\modelNameExtended}{Multimodal Attention Resilient to Incomplete datA}

\begin{document}

\begin{frontmatter}



\title{\modelName: a Multimodal Transformer Model for Incomplete Healthcare Data}

\author[UCBM]{Camillo Maria Caruso} 
\ead{camillomaria.caruso@unicampus.it}

\author[UCBM,UMU]{Paolo Soda\corref{cor1}} 
\ead{p.soda@unicampus.it, paolo.soda@umu.se}
\cortext[cor1]{Corresponding author: p.soda@unicampus.it, paolo.soda@umu.se}

\author[UCBM]{Valerio Guarrasi} 
\ead{valerio.guarrasi@unicampus.it}

  
\affiliation[UCBM]{organization={Research Unit of Computer Systems and Bioinformatics, Department of Engineering, Università Campus Bio-Medico di Roma},
            city={Roma},
            state={Italy},
            country={Europe}}

\affiliation[UMU]{organization={Department of Diagnostics and Intervention, Radiation Physics, Biomedical Engineering, Umeå University},
            city={Umeå},
            state={Sweden},
            country={Europe}}

\begin{abstract}
In healthcare, the integration of multimodal data is pivotal for developing comprehensive diagnostic and predictive models.
However, managing missing data remains a significant challenge in real-world applications.
We introduce \modelName~(\modelNameExtended), a novel transformer-based deep learning model designed to address these challenges through an intermediate fusion strategy.
Unlike conventional approaches that depend on imputation, \modelName~utilizes a modified masked self-attention mechanism, which processes only the available data without generating synthetic values.
This approach enables it to effectively handle incomplete datasets, enhancing robustness and minimizing biases introduced by imputation methods.
We evaluated \modelName~against $10$ state-of-the-art machine learning and deep learning models across $8$ diagnostic and prognostic tasks.
The results demonstrate that \modelName~outperforms existing methods in terms of performance and resilience to varying levels of data incompleteness, underscoring its potential for critical healthcare applications.
\end{abstract}



\end{frontmatter}



\section{Introduction}\label{sec:intro} 

In recent years, multimodal learning has emerged as a powerful approach for leveraging diverse data sources to achieve a comprehensive understanding of complex systems.
This is particularly relevant in domains such as healthcare, where integrating multiple data modalities, such as clinical assessments, imaging, laboratory tests, and patient histories, can significantly enhance diagnostic accuracy and treatment outcomes.
The human experience itself exemplifies multimodality, as it relies on diverse sensory inputs to form a unified perception of the environment.
Similarly, deep learning (DL) models have been developed to synthesize and analyze disparate data sources, thereby enhancing their predictive capabilities and enabling more informed decision-making in multifaceted domains such as healthcare.

Despite the promise of multimodal learning, integrating multiple data sources presents unique challenges, with data incompleteness being one of the most significant.
Missing data is a common feature of real-world datasets, arising from issues such as sensor failures, patient non-compliance, technical limitations during data collection, or privacy restrictions.
Whether the missing information relates to features within a modality or the complete absence of a modality, such gaps can severely degrade the performance of machine learning models unless effectively addressed.
Thus, the development of multimodal learning models resilient to incomplete data is critical to ensuring reliability and robustness, especially in critical fields like healthcare.

To address these challenges, multimodal fusion strategies such as early fusion, late fusion, and intermediate fusion have been extensively studied.
Early fusion, which combines features at the raw data level into a unified representation, is straightforward but highly susceptible to the effects of missing data, as it requires the availability of all feature vectors.
Late fusion, which merges outputs from independently trained models, offers flexibility when modalities are missing but often fails to capture the intricate interactions across modalities.
Intermediate fusion strikes a balance by integrating modality-specific features after initial processing. 
This forms a shared representation that enhances the ability to capture cross-modal dependencies, ultimately improving performance~\cite{bib:joint_review}. 
Therefore, especially in healthcare, it is essential to develop methods that leverage the potential of intermediate fusion while maintaining robustness to missing data.

The \modelName~(\modelNameExtended) model introduced in this work is designed to address the challenges of incomplete multimodal data.
By employing an intermediate fusion strategy, \modelName~combines modality-specific encoders with a shared attention-based encoder to effectively manage missing data.
Unlike traditional methods that rely on data imputation to fill in missing entries, \modelName~focuses exclusively on the available features, utilizing a modified  masked self-attention mechanism to process observed information only, while completely masking out the contribution of missing information. 
This approach enhances both robustness and accuracy, while reducing biases typically introduced by imputation techniques.

The manuscript is organized as follows: Section~\ref{sec:SOTA} reviews related work on multimodal learning and data handling methods; 
Section~\ref{sec:methods} introduces the \modelName~model and its architecture; 
Section~\ref{sec:exp_conf} explains the experimental setup and evaluation methodology; 
Section~\ref{sec:results} presents the obtained results, comparing \modelName~with other models under various missing data conditions; 
Section~\ref{sec:conclusion} summarizes the key findings and suggests directions for future research.

\section{State-of-the-Art}\label{sec:SOTA}

Multimodal learning combines information from diverse data sources to achieve a more comprehensive understanding of complex systems.
This mirrors the inherently multimodal nature of human perception: we rely on multiple senses, e.g., sight, sound, and touch, to develop a complete understanding of our environment.
Similarly, DL models must be designed to integrate diverse data sources to better comprehend intricate systems.
This is particularly relevant in healthcare, where clinicians utilize multimodal data, including patient histories, imaging, laboratory results, and physical examinations, to make informed decisions.
By effectively integrating such diverse information, multimodal learning models can enhance decision-making and predictive accuracy, leading to improved diagnostic outcomes and more effective treatment plans~\cite{bib:multimodal_medical}.

However, one of the primary challenges in multimodal learning is handling missing data, which frequently arises due to factors such as sensor failures, survey non-responses, or technical issues during data collection~\cite{bib:SMIL}.
Effectively managing missing data, whether it involves incomplete features within a modality or entirely absent modalities, is critical to ensuring the reliability and robustness of multimodal models.

Multimodal fusion techniques play a vital role in successfully integrating diverse data sources.
These techniques are typically categorized into three main strategies: early fusion, late fusion, and intermediate fusion (\figurename~\ref{fig:fusions}).
Each approach has distinct characteristics, making it suitable for different scenarios~\cite{bib:multimodal_medical, bib:multimodal_medical2}.

Early fusion integrates features at the raw data level, combining them into a unified representation before any significant processing.
Late fusion merges outputs from independently trained models at the decision level, offering flexibility when dealing with missing modalities.
Intermediate fusion takes a balanced approach, integrating modality-specific features after initial processing to create a shared representation.
Each of these fusion strategies has specific advantages and limitations in terms of performance, computational complexity, and their ability to manage missing data.
This is particularly significant in healthcare, where data quality and completeness often vary.
The subsequent sections provide a detailed analysis of these fusion techniques, focusing on their applications and limitations in healthcare scenarios.


\subsection{Early Fusion}
\begin{figure*}[!ht]
    \centering
    \resizebox{\textwidth}{!}{
    \begin{tikzpicture}
        \node[label={[yshift=-.1cm]below:(a)}] (early) {\includegraphics[height=6cm]{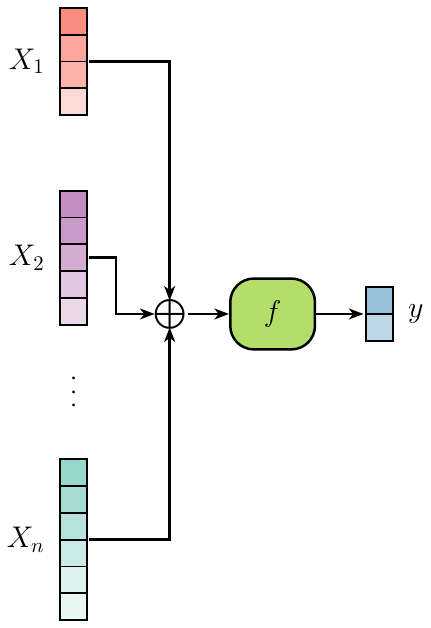}};
        \node[label={[yshift=-.1cm]below:(b)}] (joint) [right=.5cm of early]{\includegraphics[height=6cm]{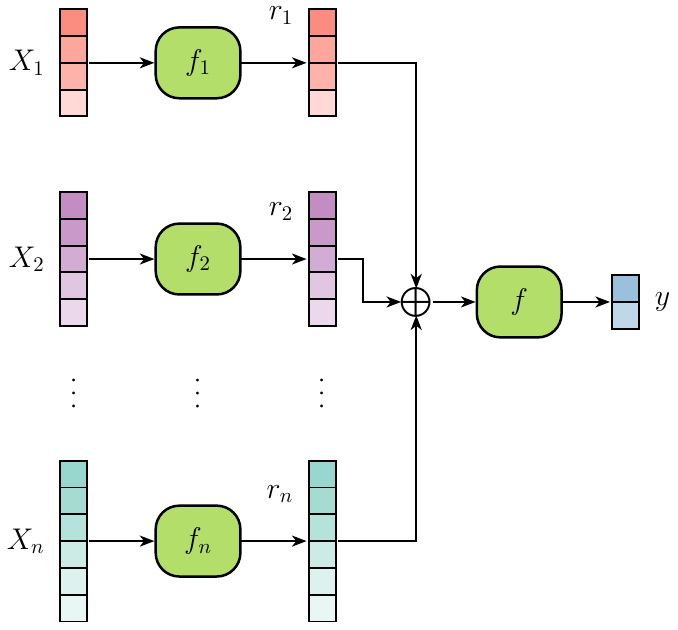}};
        \node[label={[yshift=-.1cm]below:(c)}] (late) [right=.5cm of joint]{\includegraphics[height=6cm]{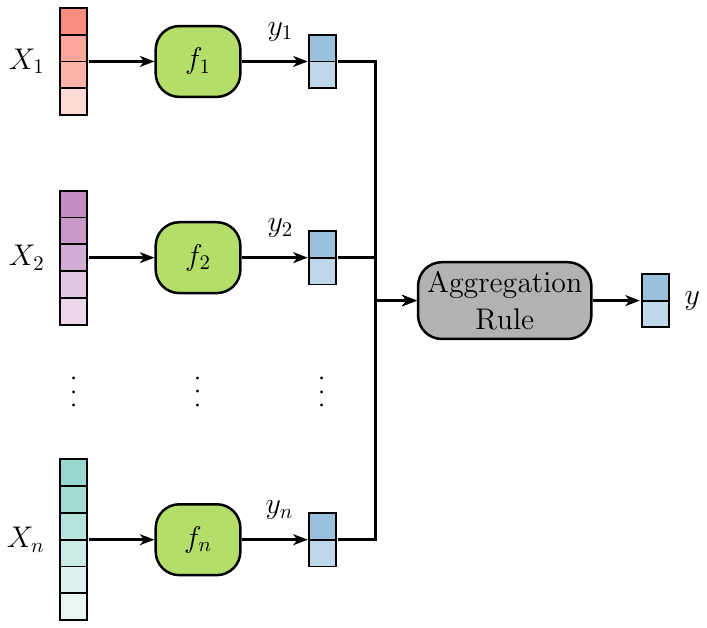}};
        \draw[-] ($(early.north)!0.4!(joint.north)$) -- ($(early.south)!0.4!(joint.south)$);
        \draw[-] ($(joint.north)!0.5!(late.north)$) -- ($(joint.south)!0.5!(late.south)$);
    \end{tikzpicture}}
    \caption{Overview of multimodal fusion strategies in DL: (a) Early fusion; (b) Intermediate fusion; (c) Late fusion. $X_1,~\ldots,~X_n$ represent the modalities; $f_{(i)}$ is a generic function representing a module or an entire model; $r_1,~\ldots,~r_n$ stand for the latent representations of the modalities; $y_{(i)}$ indicates an output of a model; $\oplus$ represent a fusion, e.g. concatenation or average, of the inputs.}
    \label{fig:fusions}
\end{figure*}

Early fusion, as illustrated in \figurename~\ref{fig:fusions}.a, involves integrating multiple data modalities at the feature level.
In this approach, the raw features $X_i$ from each modality are concatenated ($\oplus$ in the figure) to form a single feature vector, which is then fed into the learning model.
This method facilitates the early combination of information from all available modalities, making it particularly advantageous when different data sources are highly complementary.
Early fusion is conceptually straightforward and often enables the learning model to effectively exploit cross-modal correlations~\cite{bib:multimodal_medical, bib:multimodal_medical2}.

However, early fusion presents several challenges, especially when dealing with incomplete data.
Because this approach relies on the availability of all feature vectors, missing data from even a single modality can significantly degrade model performance.
Imputation is a common strategy for addressing these gaps, but it introduces potential risks such as bias and information loss~\cite{bib:imputation_review}.
Additionally, early fusion typically requires extensive preprocessing to harmonize features from different modalities, which often vary in scale and distribution.

In healthcare, early fusion can be particularly beneficial when modalities are guaranteed to be complete or when missing data is minimal and can be addressed through robust preprocessing techniques.
However, given the variability and incompleteness commonly encountered in real-world medical datasets, early fusion may struggle to perform effectively without sophisticated data-handling strategies~\cite{bib:multimodal_medical, bib:multimodal_medical2}.

\subsection{Late Fusion}
Late fusion, in contrast to early fusion, integrates modalities at the decision level, as illustrated in \figurename~\ref{fig:fusions}.c.
In this approach, separate models $f_i$ are trained for each modality $X_i$, and their predictions $y_i$ are subsequently aggregated using a predefined rule to generate the final output $y$.
This method is particularly advantageous when the modalities differ significantly in data type or exhibit varying levels of reliability.
By training separate models, late fusion allows each modality to be optimally utilized before combining their outputs.

One of the primary advantages of late fusion is its flexibility in handling missing modalities.
Since each model operates independently, missing data from one modality does not prevent predictions from being made using the available modalities.
However, late fusion has a notable limitation: it fails to fully exploit cross-modal interactions.
Because modalities are only integrated at the decision level using a static, predefined rule, rather than a dynamically learned process, potentially rich correlations between features from different data sources may remain untapped.
This drawback makes late fusion less suitable for tasks that require deep integration of modality-specific features, particularly in scenarios demanding high levels of diagnostic precision.


\subsection{Intermediate Fusion} 
Intermediate fusion, also known as joint or hybrid fusion, strikes a balance between early and late fusion.
This approach combines modality-specific features at an intermediate stage of the learning process (\figurename~\ref{fig:fusions}.b), typically after each modality has undergone initial, independent processing.
In this setup, modality-specific modules $f_i$ generate latent representations $r_i$ for each modality. 
These representations are then fused using a defined technique (denoted by $\oplus$ in \figurename~\ref{fig:fusions}.b) to form a shared representation $r_sh$. 
Finally, this shared representation is processed by a final module $f$ to produce the desired output $y$. 
This approach facilitates a richer integration of features, retaining modality-specific information while capturing inter-modal relationships during the feature extraction phase.

One of the major advantages of intermediate fusion is its ability to handle incomplete data more flexibly and effectively.
Various techniques are available for fusing latent representations from different modalities, and we recommend readers refer to the review~\cite{bib:joint_review} for an in-depth exploration of these methods.
However, intermediate fusion comes with certain challenges, including increased computational complexity and training difficulty.
The model must learn both unimodal and multimodal representations simultaneously, requiring significant computational and data resources.
These demands can pose a barrier in resource-constrained environments, such as many healthcare settings.

Despite these challenges, the dynamic nature of intermediate fusion offers significant advantages.
By allowing the model to learn how to fuse information from different sources dynamically, it enhances robustness and adaptability.
This integration of complementary information from multiple modalities enables the model to leverage the strengths of each data source while mitigating their individual weaknesses.
Such adaptability is particularly valuable in real-world healthcare scenarios, where the quality and availability of data often vary across modalities.
Intermediate fusion’s ability to handle partially missing or noisy modalities can result in more reliable predictions.
Moreover, the shared representation created through intermediate fusion fosters a deeper understanding of correlations between different data types, which is critical for complex tasks such as medical diagnosis and prognosis~\cite{bib:multimodal_medical, bib:multimodal_medical2}.

Despite its resource demands, intermediate fusion represents a promising direction for the development of DL models that are both comprehensive and resilient.
This makes it a powerful approach for enhancing decision-making in healthcare environments~\cite{bib:joint_review, bib:multimodal_medical, bib:multimodal_medical2}.

\subsection{Handling Incomplete Data}\label{sec:missing_modalities}
Real-world multimodal data are often imperfect due to missing features or modalities.
Therefore, there is a pressing need for multimodal models robust in the presence of incomplete data.
Missing data, whether involving individual features or entire modalities, is a common challenge across various fields and is often caused by issues, such as human error, survey non-responses, data corruption, or systematic loss.
Traditional approaches to address missing data typically rely on imputation techniques, which attempt to fill these gaps but can introduce biases or fail to capture underlying complexities.
For instance, we employed the k-Nearest Neighbors (kNN) imputer, which has demonstrated effectiveness in handling missing values in tabular data~\cite{bib:imputation_review, bib:review_missing, bib:NAIM}.
Additionally, we tested the Missing in Attributes (MIA) strategy, used by tree-based models to dynamically manage missing features without requiring imputation.

Healthcare settings are particularly vulnerable to the problem of incomplete data, as patients may follow different treatment plans, discontinue care for reasons such as transferring facilities, voluntarily ceasing treatment, or even passing away.
Moreover, privacy concerns further exacerbate data incompleteness in these settings~\cite{bib:review_missing}.
Many methods simply exclude patients with missing values, which can significantly reduce data availability and compromise the robustness of analyses.
Other approaches often involve imputing missing information using data from available modalities for the same subject or from other patients with similar characteristics.

Several advanced methodologies have been proposed to address the issue of missing data in multimodal contexts:

The Contrastive Masked-Attention Model integrates a Generative Adversarial Network (GAN)-based augmentation mechanism to synthesize data for missing modalities and employs contrastive learning to enhance cross-modal representations.
Masked attention ensures that only interactions between observed modalities are captured, thereby minimizing the introduction of extraneous noise~\cite{bib:COM}.

The Cascaded Multi-Modal Mixing Transformers implement a cascaded cross-attention architecture to effectively integrate multiple available modalities, enabling robust performance even when some modalities are missing.
This approach offers flexibility and adaptability in fusing modality-specific information~\cite{bib:cascaded}.

The Missing Modalities in Multimodal healthcare framework employs task-guided, modality-adaptive similarity metrics to identify similar patients and impute missing data.
By leveraging auxiliary information from comparable patients, this method preserves the underlying relationships in multimodal healthcare data~\cite{bib:m3care}.

Shared-Specific Feature Modeling disentangles shared features from modality-specific ones, enabling efficient handling of missing data during both training and inference.
By learning shared features across all available modalities, this approach ensures the retention of essential representations while maintaining model performance~\cite{bib:shaspec}.

The Severely Missing Modality model uses a Bayesian meta-learning framework to approximate latent representations for missing modalities.
This method is designed to handle incomplete data during both training and testing, offering robust generalization capabilities even when data availability is severely limited~\cite{bib:SMIL}.

These methodologies highlight diverse strategies for compensating for missing information, including identifying shared latent representations, generating synthetic data, and leveraging auxiliary patient information.
By reducing dependence on complete multimodal datasets, these approaches improve the practicality of multimodal models in real-world clinical and resource-constrained settings.
However, both traditional and DL-based approaches share a common limitation: they rely on artificially filling data gaps, which can introduce bias and compromise task accuracy.

To address this limitation, we propose a model that exclusively utilizes the available features and modalities, avoiding the generation of synthetic data.
By focusing solely on effectively leveraging observed information, our approach aims to enhance robustness and reliability, even in scenarios with severely missing data.


\section{Methods}\label{sec:methods}
In this work, we propose \modelName~(\modelNameExtended), a multimodal model specifically designed to address the challenges of incomplete features and modalities in multimodal healthcare data.
The model effectively integrates data modalities that may be incomplete or entirely absent, offering a robust solution for predictive analysis without relying on traditional data imputation techniques or synthetic data generation.

This section first provides an overview of the \modelName~model design.
We then focus on the architecture of the modality-specific encoders, the strategies employed for handling missing data, and the regularization techniques implemented during training to enhance generalizability under incomplete input conditions.

\subsection{Model}

\begin{figure*}[!t]
    \centering
    \includegraphics[width=10cm]{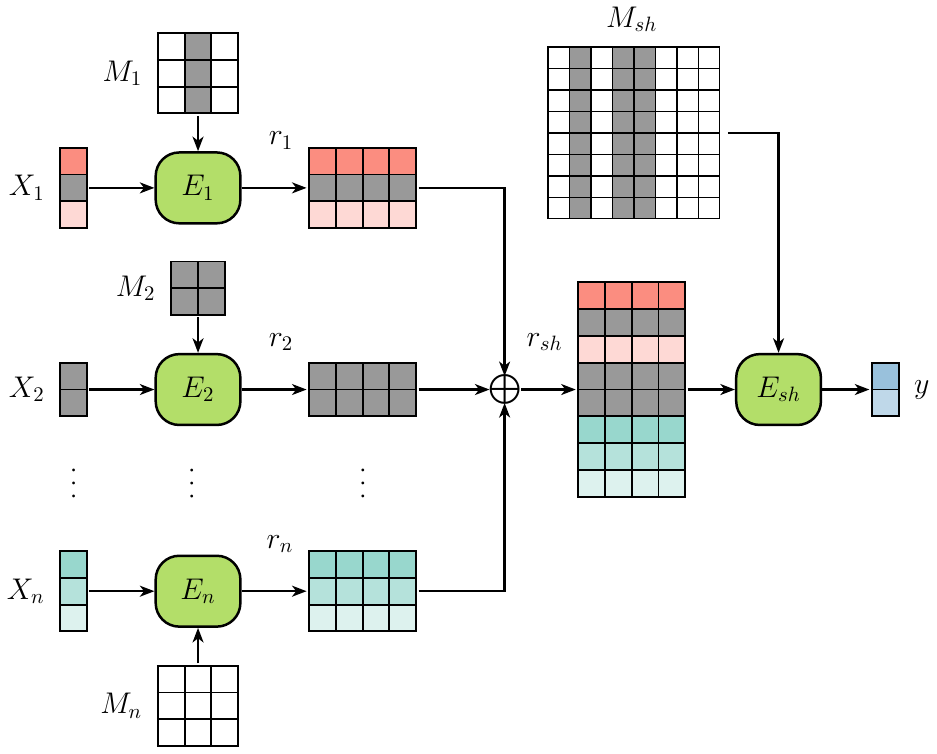}
    \caption{\modelName~architecture. Each modality-specific encoder $E_i$ takes the modality $X_i$ as input to generate the latent representation $r_i$. Then, the shared encoder $E_{sh}$ elaborates the concatenation of the latent representations $r_{sh}$ to get the final output $y$. In the figure, a gray square represents a missing feature or its respective element in the masking matrix.}
    \label{fig:model}
\end{figure*}

\modelName~is specifically designed to be resilient to incomplete data and modalities without relying on imputation.
It employs intermediate fusion, using modality-specific encoders and a shared encoder with a masked self-attention mechanism to combine latent representations while effectively managing missing data.
The architecture incorporates multiple modality-specific encoders for each data modality.
In this work, we focus on multimodal problems where the modalities represent tabular data describing various aspects of a patient’s condition.
Thus, \modelName~utilizes separate NAIM~\cite{bib:NAIM} modules as modality-specific encoders (\figurename~\ref{fig:model}).
These encoders integrate a modified  masked multi-head attention mechanism that selectively focuses on available features within each modality while completely ignoring missing ones.
This mechanism ensures robustness by excluding absent features from attention computations.

Each tabular modality $X_i$, where $i=1,\ldots,n$, is encoded into embeddings using look-up tables~\cite{bib:NAIM}, which represent missing entries with a specific non-trainable embedding.
The modality embeddings are then processed by their corresponding encoders $E_i$, which compute query, key, and value matrices, denoted as $Q_i$, $K_i$, and $V_i$, using linear transformations:

\begin{equation}\label{eq:embeddings}
    \left\{
    \begin{aligned}
    Q_i &= X_i \cdot W^Q_i &&&d^h=d^e/h \\
    K_i &= X_i \cdot W^K_i &&&W^Q_i, W^K_i, W^V_i \in \mathbbm{R}^{d^e \times d^h} \\
    V_i &= X_i \cdot W^V_i &&&Q_i, K_i, V_i \in \mathbbm{R}^{|X_i| \times d^h} 
    \end{aligned}
    \right.
\end{equation}
where $W^Q_i$, $W^K_i$ and $W^V_i$ are learnable weight matrices. 
These transformations reduce dimensionality to $d^h$, determined by the token dimensions $d^e$ and the number of heads $h$ in the model.
Next, a modified masked self-attention mechanism is applied:
\begin{equation}\label{eq:new_attn}
    \mathit{MSA}(Q_i,K_i,V_i) = \mathit{ReLU} \left( \mathit{softmax} \left(\frac{Q_iK_i^T}{\sqrt{d^h}} + M_i\right) + M_i^T \right)V_i
\end{equation}
where the masking matrix $M_i$ ensures that missing features do not influence the latent representation $r_i$.
The elements of $M_i$ are defined as follows:
\begin{equation}\label{eq:pad} 
\resizebox{.9\columnwidth}{!}{$
\begin{aligned} 
    M_i^{kj} = \left\{ 
    \begin{aligned} 
        -&\infty &\text{if}~X_i^j\text{ is missing}\\
        &0 &\text{otherwise}
    \end{aligned}
    \right.,
    \quad
    &M_i = 
    \renewcommand\arraystretch{1}\begin{bmatrix}[*5{C{5.6mm}}]
                0 &             -$\infty$ &     ... &           0 &     0 \\
                0 &             -$\infty$ &     ... &           0 &     0 \\
              \rotatebox{90}{...} &           \rotatebox{90}{...} &     \large\rotatebox{135}{...} &         \rotatebox{90}{...} &           \rotatebox{90}{...} \\
                0 &             -$\infty$ &     ... &           0 &     0 \\
                0 &             -$\infty$ &     ... &           0 &     0 \\
                \end{bmatrix}
\end{aligned}$}
\end{equation}
This operation effectively zeroes out weights associated with missing features after applying $\mathit{softmax}$ and $\mathit{ReLU}$.
Each modality-specific encoder $E_i$ generates a latent representation $r_i$, of dimensions $|X_i| \times d^e$, where $|X_i|$ represents the number of tokens in the modality, e.g., the number of features in the $i-$th modality.
These latent representations are then concatenated to form a joint representation $r_{sh}$, composed only of available information, with null vectors representing missing features.
This multimodal representation is passed to the shared encoder $E_{sh}$, which computes its own query, key, and value matrices, denoted as $Q_{sh}$, $K_{sh}$, and $V_{sh}$, as follows:
\begin{equation}\label{eq:embeddings_sh}
    \left\{
    \begin{aligned}
    Q_{sh} &= r_{sh} \cdot W^Q_{sh} &&&W^Q_{sh}, W^K_{sh}, W^V_{sh} \in \mathbbm{R}^{d^e \times d^h} \\ 
    K_{sh} &= r_{sh} \cdot W^K_{sh} &&&Q_{sh}, K_{sh}, V_{sh} \in \mathbbm{R}^{\sum_i |X_i| \times d^h} \\
    V_{sh} &= r_{sh} \cdot W^V_{sh} 
    \end{aligned}
    \right.
\end{equation}
where $W^Q_{sh}$, $W^K_{sh}$ and $W^V_{sh}$ are weights matrices learned during training. 
As with the modality-specific encoders, dimensionality is reduced to $d^h$ through linear transformations.

A similar modified masked self-attention mechanism is then applied:
\begin{equation}\label{eq:new_attn_sh}
\resizebox{.9\columnwidth}{!}{$
    \mathit{MSA}(Q_{sh},K_{sh},V_{sh}) = \mathit{ReLU} \left( \mathit{softmax} \left(\frac{Q_{sh}K_{sh}^T}{\sqrt{d^h}} + M_{sh}\right) + M_{sh}^T \right)V_{sh} 
$}
\end{equation}
where the masking matrix $M_{sh}$ ensures that missing modalities do not impact the final shared representation.
This matrix operates in the same manner as $M_i$, zeroing out weights associated with missing modalities.
The elements of the masking matrix are defined as follows:
\begin{equation}\label{eq:pad_sh} 
\resizebox{.9\columnwidth}{!}{$
\begin{aligned} 
    M_{sh}^{kj} = \left\{ 
    \begin{aligned} 
        -&\infty &\text{if}~r_{sh}^j\text{ is missing}\\
        &0 &\text{otherwise}
    \end{aligned}
    \right.,
    \quad
    &M_{sh} = 
    \renewcommand\arraystretch{1}\begin{bmatrix}[*5{C{5.6mm}}]
                0 &             -$\infty$ &     ... &           0 &     0 \\
                0 &             -$\infty$ &     ... &           0 &     0 \\
              \rotatebox{90}{...} &           \rotatebox{90}{...} &     \large\rotatebox{135}{...} &         \rotatebox{90}{...} &           \rotatebox{90}{...} \\
                0 &             -$\infty$ &     ... &           0 &     0 \\
                0 &             -$\infty$ &     ... &           0 &     0 \\
                \end{bmatrix}
\end{aligned}$}
\end{equation}
$M_{sh}$ sums the $-\infty$ values to weights that need to be ignored, effectively zeroing them after applying the $\mathit{softmax}$ and $\mathit{ReLU}$ operations.

Finally, the joint representation is passed through a fully connected layer, which predicts the output $y$.
The training process minimizes prediction error, updating the weights of both the modality-specific encoders ($E_i$) and the shared encoder ($E_{sh}$) via end-to-end backpropagation.

This architecture qualifies \modelName~as an intermediate fusion model, performing fusion at the latent representation level.
By dynamically optimizing all encoders, \modelName~balances the contributions of different modalities and adapts to maximize their utility during training and inference.
The use of a masked multi-head attention mechanism ensures the model focuses adaptively on informative parts of the input, completely ignoring missing data.
This approach allows each modality to contribute based on its completeness, resulting in accurate and reliable multimodal representations.



\subsection{Regularization Technique for Missing Data}

To enhance the model's generalizability under incomplete input conditions, we employ regularization strategies during training that improve its resilience to varying degrees of data incompleteness.
These strategies ensure that even when some modalities or features are unavailable, the model can still generate accurate and meaningful outputs~\cite{bib:NAIM, bib:cascaded}.
During training, we simulate a relaxed missing data setting where each modality or feature is treated as potentially missing, while maintaining at least one available data point per patient.
This approach allows the model to learn how to handle different levels of missingness effectively, making it particularly well-suited to the variability typically found in clinical datasets.
By encouraging the model to extract meaningful representations from each available modality, these masking strategies promote robustness against incomplete information during both training and inference.

\paragraph{Modality Dropout}
During training, the model uses a stochastic masking procedure to simulate incomplete data scenarios.
Given a sample $X=\{X_1, \ldots, X_n\}$, where $n$ represents the number of modalities, let $v_m \leq n$ denote the number of non-missing modalities in the sample (where $v_m=n$ for fully populated data or $v_m<n$ for partially missing data).
A binary decision variable determines whether masking will be applied to the sample $X$.
If the sample is selected for masking, a random count $c_m$ of modalities to mask is chosen uniformly from the set $\{1, 2, \ldots, v_m-1\}$, ensuring that at least one modality remains unmasked.
Finally, $c_m$ non-missing modalities are randomly selected, and their values are set to \textit{missing}, producing the augmented sample.

\paragraph{Feature Dropout}
Similarly, when a tabular modality $X_i$ is set as present, a similar stochastic masking procedure is applied at the feature level.
A binary decision variable determines whether masking will be applied to the features of the modality $X_i$.
If masking is applied, a random count $c_i$ of features to mask is chosen uniformly from the set $\{1, 2, \ldots, v_i-1\}$, where $v_i$ is the number of non-missing features of the modality $X_i$. 
This ensures that at least one feature remains unmasked.
Finally, $c_i$ non-missing features within $X_i$ are randomly chosen and set to \textit{missing}, resulting in the augmented modality. 


\section{Experimental Configuration}\label{sec:exp_conf}
In this section, we first describe the datasets used in the experiments and the preprocessing applied (Section~\ref{sec:data}).
We then outline the combinations of models and imputers employed as competitors (Section~\ref{sec:competitors}).
Finally, we present the metrics used for evaluation (Section~\ref{sec:metrics}).

\subsection{Data}\label{sec:data}
We evaluated \modelName~and its competitor models on two publicly available datasets across eight diagnostic and prognostic tasks (details in \tablename~\ref{tab:data}).
These tabular datasets represent real-world scenarios where patient data is often incomplete, highlighting the need for methods that are resilient to missing information.

\begin{table*}[!ht]
    \centering
    \resizebox{\textwidth}{!}{
    \begin{tabular}{c|c|c|cc|c}
         \toprule
         \textbf{Dataset} & \textbf{Task} & \textbf{\# of samples} & \multicolumn{2}{c|}{\textbf{Class distribution}} &     \makecell{\textbf{Modalities Info}\\ \textbf{(\% of missing features and modalities)}} \\ 
         \midrule
         \multirow{21}{*}{\textbf{ADNI}~\cite{bib:ADNI}}  & \makecell{Diagnosis\\ Binary} & $953$ & $\mathbf{CN}$: $542$ & $\mathbf{AD}$: $411$ & \makecell{Assessment: 37 (f: 35\% - m: 0\%)\\ Biospecimen: 47 (f: 57\% - m: 5\%)\\ Image Analysis: 14 (f: 38\% - m: 1\%)\\ Subject Characteristics: 17 (f: 37\% - m: 0\%)} \\\cline{2-6}
         & \makecell{Diagnosis\\ Multiclass} & $2066$ & \makecell{$\mathbf{CN}$: $542~$\\ $\mathbf{LMCI}$: $690$} & \makecell{$\mathbf{EMCI}$: $423$\\ $\mathbf{AD}$: $411$} & \makecell{Assessment: 37 (f: 32\% - m: 0\%)\\ Biospecimen: 47 (f: 56\% - m: 7\%)\\ Image Analysis: 14 (f: 36\% - m: 1\%)\\ Subject Characteristics: 17 (f: 38\% - m: 0\%)} \\\cline{2-6} 
         & \makecell{Prognosis\\ 12 months} & $1340$ & \makecell{$\mathbf{CN}$: $427$\\ $\mathbf{MCI}$: $797$} & $\mathbf{Dementia}$: $116$ & \makecell{Assessment: 126 (f: 33\% - m: 0\%)\\ Biospecimen: 51 (f: 47\% - m: 0\%)\\ Image Analysis: 18 (f: 27\% - m: 0\%)\\ Subject Characteristics: 21 (f: 29\% - m: 0\%)} \\\cline{2-6} 
         & \makecell{Prognosis\\ 24 months} & $1159$ & \makecell{$\mathbf{CN}$: $428$\\ $\mathbf{MCI}$: $535$} & $\mathbf{Dementia}$: $196$ & \makecell{Assessment: 126 (f: 34\% - m: 0\%)\\ Biospecimen: 51 (f: 46\% - m: 0\%)\\ Image Analysis: 18 (f: 27\% - m: 0\%)\\ Subject Characteristics: 21 (f: 28\% - m: 0\%)} \\\cline{2-6} 
         & \makecell{Prognosis\\ 36 months} & $856$ & \makecell{$\mathbf{CN}$: $239$\\ $\mathbf{MCI}$: $420$} & $\mathbf{Dementia}$: $197$ & \makecell{Assessment: 126 (f: 27\% - m: 0\%)\\ Biospecimen: 51 (f: 37\% - m: 0\%)\\ Image Analysis: 18 (f: 27\% - m: 0\%)\\ Subject Characteristics: 21 (f: 23\% - m: 0\%)} \\\cline{2-6}
         & \makecell{Prognosis\\ 48 months} & $693$ & \makecell{$\mathbf{CN}$: $269$\\ $\mathbf{MCI}$: $280$} & $\mathbf{Dementia}$: $144$ & \makecell{Assessment: 126 (f: 36\% - m: 0\%)\\ Biospecimen: 51 (f: 47\% - m: 0\%)\\ Image Analysis: 18 (f: 28\% - m: 0\%)\\ Subject Characteristics: 21 (f: 30\% - m: 0\%)} \\\midrule
         \multirow{4}{*}{\textbf{AIforCOVID}~\cite{bib:AI4COVID}}  & Mild/Severe & $1585$ & $\mathbf{Mild}$: $839$ & $\mathbf{Severe}$: $746$ & \makecell{Blood Analysis: 14 (f: 31\% - m: 1\%)} \\\cline{2-5}
         & Death    & $1585$ & $\mathbf{Censored}$: $1336$ & $\mathbf{Uncensored}$: $249$ & \makecell{Clinical History: 13 (f: 24\% - m: 9\%)\\ Personal Info: 2 (f: 0\% - m: 0\%)\\ Admission State: 5 (f: 8\% - m: 0\%)} \\
         \bottomrule
    \end{tabular}}
    \caption{Datasets' details consisting of: dataset name and reference, the task name, the number of samples, the classes' distribution, and the different tabular modalities, with the corresponding number of features and the respective rates of original missing features (f) and modalities (m).}
    \label{tab:data}
\end{table*}

The first dataset was obtained from the Alzheimer’s Disease Neuroimaging Initiative (ADNI) database (adni.loni.usc.edu)~\cite{bib:ADNI}. 
The ADNI was launched in 2003 as a public-private partnership, led by Principal Investigator Michael W. Weiner, MD. The primary goal of ADNI has been to test whether serial magnetic resonance imaging, positron emission tomography, other biological markers, and clinical and neuropsychological assessment can be combined to measure the progression of mild cognitive impairment (MCI) and early Alzheimer’s disease (AD), with respect to the Cognitively Normal (CN) group.
For our study, we used tabular data derived from four baseline modalities: Assessment (cognitive and neuropsychological scores), Biospecimen (CSF, ApoE genotyping, and lab data), Image analysis (MRI and PET neuroimaging biomarkers), and Subject Characteristics (family history, demographics).
We analyzed data from ADNI 1, GO, 2, and 3 phases.
Diagnostic tasks included binary classification (CN vs. AD) and ternary classification (CN vs. AD vs. MCI), reflecting real clinical differentiation scenarios.
Additionally, prognostic tasks aimed to predict whether treatment intervention might be necessary at specific future points (12, 24, 36, and 48 months post-recruitment).
These tasks classified patients into CN, MCI, and Dementia categories.
\tablename~\ref{tab:data} provides details on patient distributions for both baseline and follow-up classifications.

The second dataset, AIforCOVID~\cite{bib:AI4COVID}, contains clinical data from six Italian hospitals, collected during the first wave of the COVID-19 pandemic (March–June 2020).
Data was recorded at the time of hospitalization of symptomatic patients and subsequently anonymized and reviewed.
All patients tested positive for SARS-CoV-2 via RT-PCR, with 5\% confirmed only after a second test.
Each patient was classified as either mild (discharged or hospitalized without ventilatory support) or severe (requiring non-invasive ventilation, ICU care, or deceased). 
Additionally, we evaluated the proposed approach on a death prediction task, classifying patients as either censored (alive) or uncensored (deceased).

\modelName's performance across diverse medical tasks provides valuable insights into its resilience and adaptability in real-world healthcare scenarios.
By leveraging the ADNI and AIforCOVID datasets, we demonstrate the model's ability to handle challenges such as missing modalities and heterogeneous data distributions.

These experiments emphasize the importance of multimodal fusion techniques that are not only robust to missing data but also capable of learning from complex, interrelated medical datasets.
Such advancements provide the foundation for more resilient and flexible DL models in healthcare, ultimately supporting clinicians in making better-informed decisions under real-world constraints.

\subsection{Competitors}\label{sec:competitors}

We conduct an extensive comparison of our methodology against early, late, and intermediate fusion approaches that use missing data imputation as a preprocessing step before model training.
Additionally, we benchmark against tree-based models that manage missing values using the Missing In Attributes (MIA) strategy.
We choose not to include generative approaches, as these are primarily developed for imaging modalities rather than tabular data, and they introduce additional complexity and randomness, which can hinder reproducibility.
Instead, we focus on interpretable, efficient approaches widely adopted in clinical settings, where model simplicity and reliability are crucial.
Our analysis includes a total of $10$ distinct competitor models, each combined with the k-Nearest Neighbors (kNN) imputation technique, as this method outperformed other imputation strategies in prior studies~\cite{bib:imputation_review, bib:NAIM, bib:review_ML}.  
For all models, we used default hyperparameters. 
Indeed, as shown in~\cite{bib:tune_param}, in line with the ``No Free Lunch'' theorem, the authors empirically observe that, in many cases, tuning hyperparameters does not lead to significantly better performance compared to the default values suggested in the literature.

\tablename~\ref{tab:competitors} provides an overview of the competitors. 
The first column categorizes models as ML or DL approaches, the second specifies the base learners, the subsequent columns indicate the techniques used for handling missing data, and the final columns describe the fusion strategies employed.
This results in 32 competitor configurations, each marked by an ``$\times$'' in the respective columns.



\begin{table}[!ht]
    \centering
    \resizebox{\columnwidth}{!}{
    \setlength\extrarowheight{1.5pt}\begin{tabular}{c|c||c|c||c|c|c}
        \toprule
        & & \multicolumn{2}{c||}{\textbf{Imputer}}  & \multicolumn{3}{c}{\textbf{Multimodal Strategies}}\\
         \textbf{Type} & \textbf{Model} & ~With~ & Without & Early & Intermediate & Late \\
         \midrule
         \multirow{6}{*}{\rotatebox[origin=c]{90}{\makebox[0pt][c]{\parbox{2cm}{\centering\textbf{Machine \\Learning}}}}} & AdaBoost & \large{$\times$} & & \large{$\times$} & & \large{$\times$}\\\cline{3-7}
         & Decision Tree & \large{$\times$} & \large{$\times$} & \large{$\times$} & & \large{$\times$} \\\cline{3-7}
         & HistGradientBoost & \large{$\times$} & \large{$\times$} & \large{$\times$} & & \large{$\times$}\\\cline{3-7}
         & Random Forest & \large{$\times$} & \large{$\times$} & \large{$\times$} & & \large{$\times$}\\\cline{3-7}
         & SVM & \large{$\times$} & & \large{$\times$} & & \large{$\times$}\\\cline{3-7}
         & XGBoost & \large{$\times$} & \large{$\times$} & \large{$\times$} & & \large{$\times$}\\
         \midrule
         \multirow{4}{*}{\rotatebox[origin=c]{90}{\makebox[0pt][c]{\parbox{2cm}{\centering\textbf{Deep \\Learning}}}}} & MLP & \large{$\times$} & & \large{$\times$} & \large{$\times$} & \large{$\times$}\\\cline{3-7}
         & TabNet & \large{$\times$} & & \large{$\times$} & \large{$\times$} & \large{$\times$}\\\cline{3-7}
         & TabTransformer & \large{$\times$} & & \large{$\times$} & \large{$\times$} & \large{$\times$}\\\cline{3-7}
         & FTTransformer & \large{$\times$} & & \large{$\times$} & \large{$\times$} & \large{$\times$}\\
         \bottomrule
    \end{tabular}}
    \caption{Combinations of models, missing techniques and fusion strategies used as competitors, represented by an ``$\times$'', in the experiments.}
    \label{tab:competitors}
\end{table}

To thoroughly evaluate the performance of our proposed methodology, we designed experiments comparing it against a diverse set of baseline models.
These experiments involve both ML models, which may rely on imputation or employ the MIA strategy, and DL models paired with imputation techniques.

The ML models include AdaBoost, Decision Trees, HistGradientBoost, Random Forests, Support Vector Machines (SVM), and XGBoost. 
AdaBoost~\cite{bib:adaboost} is a cascading ensemble model that prioritizes hard-to-classify instances, offering robustness across diverse datasets. 
Decision Trees~\cite{bib:DT} are highly interpretable models that visually represent decision-making processes, providing insights into complex data relationships.
HistGradientBoost~\cite{bib:review_ML} offers an efficient variation of gradient boosting, optimized for handling large datasets with improved speed and memory usage. 
Random Forests~\cite{bib:RF} is an ensemble of decision trees known for robustness against overfitting and enhanced reliability.
SVM~\cite{bib:SVM}, equipped with an RBF kernel, is included for its versatility in handling non-linear data separations, providing a contrast to tree-based models. 
XGBoost~\cite{bib:xgboost}, an advanced boosting model, employs a gradient descent procedure to minimize loss and is highly effective for tabular datasets.

Additionally, we evaluate DL models paired with imputation methods. 
These approaches include Multilayer Perceptron (MLP), TabNet, TabTransformer, and FTTransformer. 
MLP~\cite{bib:mlp} is a foundational DL model that captures complex relationships between features, offering a baseline for comparison.
TabNet~\cite{bib:tabnet} leverages self-attention to dynamically select features, improving interpretability and decision-making. 
TabTransformer~\cite{bib:tabtransformer} uses transformer-based self-attention mechanisms to embed categorical features and capture complex inter-feature relationships. 
FTTransformer~\cite{bib:fttransformer} further explores transformers' potential, using distinct embedding strategies for numerical and categorical features.

The selected DL models were evaluated using both early and late fusion approaches.
In addition, we developed intermediate fusion variations of these models, where the respective architectures were employed for both modality-specific encoders and shared encoder settings.
These intermediate fusion configurations assess the models' ability to concurrently handle multiple input types, leveraging the shared encoder to effectively combine information from various modalities.

These experiments were designed to comprehensively assess the strengths and limitations of each competitor model across various settings.
Our comparisons aim to benchmark the performance of our intermediate fusion methodology against both traditional ML approaches and advanced DL competitors.
By exploring a wide range of techniques, we highlight the effectiveness of our approach in handling incomplete and heterogeneous multimodal healthcare data.

\subsection{Preprocessing}\label{sec:preprocessing}
For each dataset, we normalize the numerical features using a Min-Max scaler and apply one-hot encoding to the categorical features before feeding them into the models.
However, for models such as \modelName, NAIM, HistGradientBoost, TabNet, TabTransformer, FTTransformer, and XGBoost, one-hot encoding is not applied, as their implementations can directly handle categorical features.
The preprocessing steps are calibrated using the training data and then applied to the validation and testing sets.

\subsection{Missingness Evaluation}\label{sec:eval}
Our experiments center on generating \emph{Missing Completely At Random} (MCAR) values artificially as it represents the broadest class of missing data type without the introduction of any bias. 
Our goal is to test our model under diverse missing data conditions by introducing missing values and modalities at various rates, denoted as $p$, across both the training and testing sets. 
Specifically, we generate separate missing rates for the training and test sets, set to $0\%$, $5\%$, $10\%$, $30\%$, $50\%$, and $75\%$. 
No additional missing values were introduced if the generated rate was lower than the dataset's pre-existing level of missingness, resulting in a variable number of experiments per dataset.

Moreover, we performed two types of experiments to evaluate our model under different missingness scenarios: 
\begin{itemize}
    \item \textbf{Missing Modalities}: where entire modalities for each patient within the dataset are masked, simulating scenarios where certain data sources were absent.
    \item \textbf{All Missing}: where a certain percentage of individual elements across the entire dataset are masked, thereby affecting all modalities simultaneously and eventually obtaining both missing features and modalities.
\end{itemize}

This approach allows us to explore a wide range of data completeness scenarios, starting from the dataset's original missing rate ($\Omega$) to extreme cases where up to $75\%$ of features or modalities are missing.
Given a target missing rate $p$, the total number of samples $N$ in the dataset, and the number of elements per sample (either the number of features $|X_i|$ or the number of modalities $n$), the total number of values to be masked is calculated as $N \cdot |X_i| \cdot p$ or $N \cdot n \cdot p$, respectively.
This calculation also takes into account any pre-existing missing entries $m_j$ for each sample $j$, representing either the number of missing features or missing modalities.
Thus, the adjusted number of values to be masked was computed as $N \cdot |X_i| \cdot p - \sum_j m_j$ or $N \cdot n \cdot p - \sum_j m_j$.
We then generated a random masking matrix of dimensions $N \times |X_i|$ or $N \times n$ to reflect the structure of the dataset, ensuring that at least one value in any fully masked row or column was replaced to avoid complete data loss in a specific dimension. 
As a result, the samples and features exhibited varying degrees of missingness, all conforming to the MCAR framework.

\subsection{Evaluation Metrics}\label{sec:metrics}
To evaluate the models, each dataset is divided into five stratified cross-validation splits, ensuring the original class distribution is preserved.
Within each fold, $20\%$ of the training samples is reserved for validation.
The performance of each experiment is assessed by averaging the values of the Area Under the Receiver Operating Characteristic Curve (AUC) or the Matthews Correlation Coefficient (MCC) computed across the cross-validation folds.

The AUC is a widely used metric for evaluating classification tasks, as it measures the model's ability to distinguish between positive and negative instances across all possible thresholds.
This makes it a comprehensive indicator of performance, especially for tasks with balanced or slightly imbalanced class distributions.
We employed the AUC for tasks such as the ADNI diagnostic and AIforCOVID prognostic evaluations, where the class distributions were relatively balanced.
The AUC effectively captures the trade-off between true positive and false positive rates, providing a robust measure of classifier performance.

For highly imbalanced datasets, the AUC may not reliably reflect model performance, as it can be disproportionately influenced by the majority class.
In such cases, we use the MCC, which accounts for all four categories of the confusion matrix (true positives, false positives, true negatives, and false negatives).
The MCC provides a more balanced view, making it particularly suitable for scenarios with significantly unbalanced positive and negative classes.
We employ the MCC for tasks such as the ADNI prognostic evaluations and the AIforCOVID death prediction task, where class imbalance is pronounced.

\subsection{Fusion Analysis}\label{sec:ablation}
As an additional analysis of our proposed approach, we also compared \modelName~with NAIM~\cite{bib:NAIM}, our unimodal baseline model, which serves as the foundation for our intermediate fusion methodology. 
This comparison illustrates the progression from unimodal analysis to the more sophisticated fusion mechanism that supports our proposed methodology. 
By conducting this comparative analysis, we provide deeper insights into how our approach builds on and improves upon the unimodal baseline, emphasizing the advantages and advancements made through the intermediate fusion technique.

Specifically, we evaluated our approach against the NAIM model under both early and late fusion configurations, which are representative of different strategies for leveraging multimodal information. 
In the early fusion scenario, we concatenated the features from all modalities before inputting them into the model, effectively treating all available information as a unified input space. 
This allowed us to explore the interactions between different modalities at an early stage of the modeling process. 
In the late fusion scenario, we train separate models for each modality, subsequently combining their predictions by averaging the corresponding decision profiles. 
This approach enables each modality to be independently modeled, allowing the strengths of each individual modality to contribute to the final decision in an aggregated manner.

The results of this extended comparison help demonstrate the value of our proposed fusion strategies and their ability to extract meaningful insights from multimodal data. 
By examining the performance differences between early and late fusion configurations, we better understand the unique strengths of each strategy and how our intermediate fusion approach effectively balances them.
This balance allows \modelName~to extract meaningful insights from multimodal data, achieving optimal performance by leveraging both modality-specific information and cross-modal interactions.


\section{Results}\label{sec:results}
As described in the previous section, we compare \modelName~with $32$ leading competitor models across both ML and DL approaches for tabular data.
We evaluated performance under two distinct experimental configurations: missing modalities and all missing.
Each configuration involved $36$ combinations of missing value percentages in the training and testing sets across $8$ tasks, resulting in a total of $18432$ experiments ($2304$ per task).
However, due to the pre-existing missing rates in the datasets, the final results were derived from $12192$ experiments across all tasks.

To analyze the results, we grouped the competitors into categories for clearer comparisons and visualized the average performance achieved during five-fold cross-validation.
The results are presented in eight separate plots (\figurename s~\ref{fig:MLmod}, \ref{fig:MLall}, \ref{fig:DLmod}, \ref{fig:DLall}), each corresponding to a specific task.
Each task-specific plot includes charts, one for each level of missing data in the training set, separately showing performance metrics (y-axis) as the missing rate in the testing set increased (x-axis).
Note that the number of charts and points along the x-axis may vary due to the initial missing rates ($\Omega$). 
These charts allow for a detailed comparison of \modelName~against different groups of competitors, specifically ML (\figurename s~\ref{fig:MLmod} and \ref{fig:MLall}) and DL models (\figurename s~\ref{fig:DLmod} and \ref{fig:DLall}).

\subsection{\modelName~vs. ML}

\begin{figure*}[!t]
    \centering
    \includegraphics[width=\linewidth, trim={4cm 1cm 4cm 2cm}, clip]{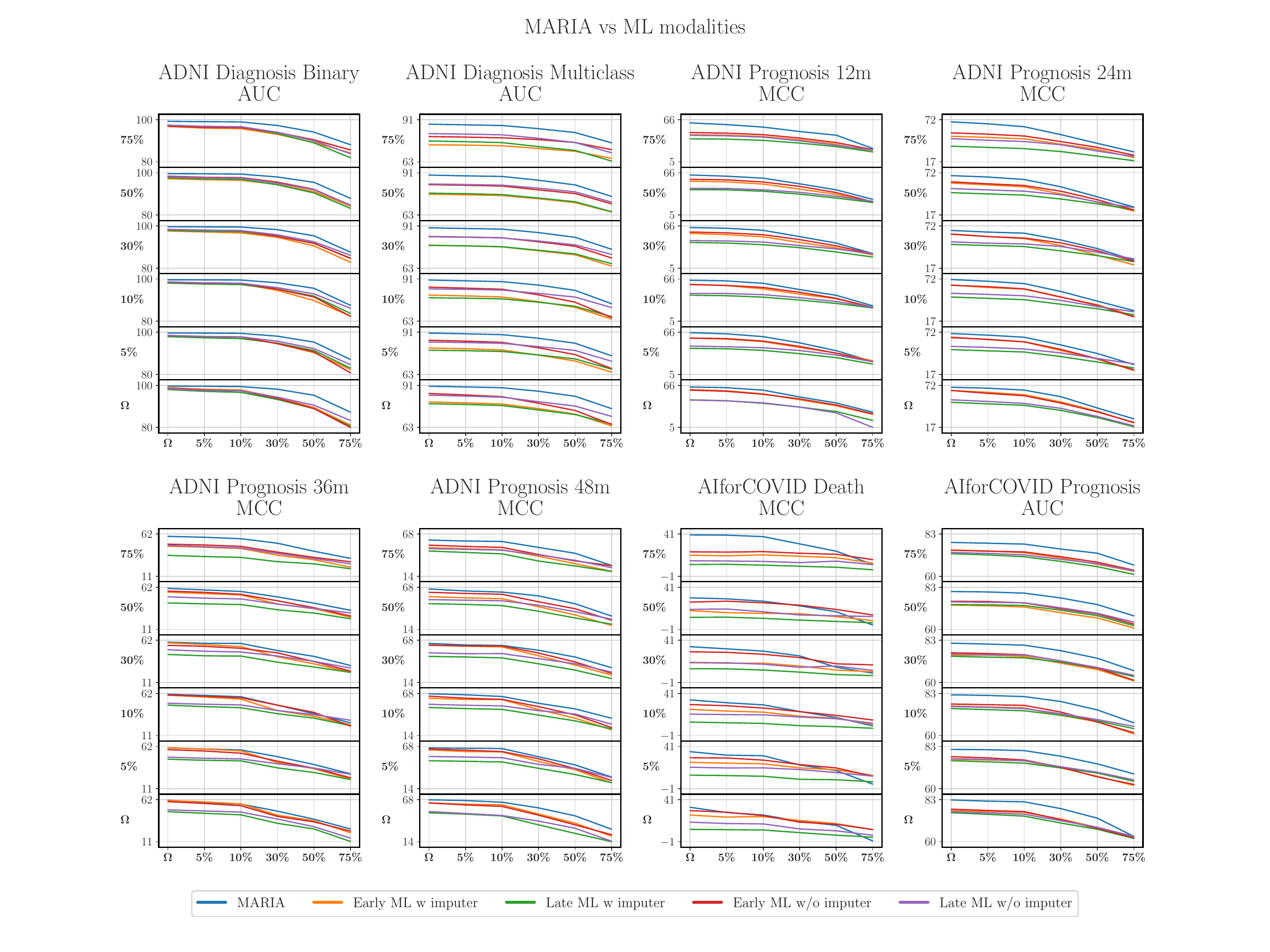}
    \caption{\modelName~vs. ML in the ``missing modalities'' scenario. Each plot, one for each task, reports different charts, one for each of the missing rates in the training set, showing how the performance (y-axis) changes as the missing rate in the test set increases.}
    \label{fig:MLmod}
\end{figure*}

\begin{figure*}[!t]
    \centering
    \includegraphics[width=\linewidth, trim={4cm 1cm 4cm 2cm}, clip]{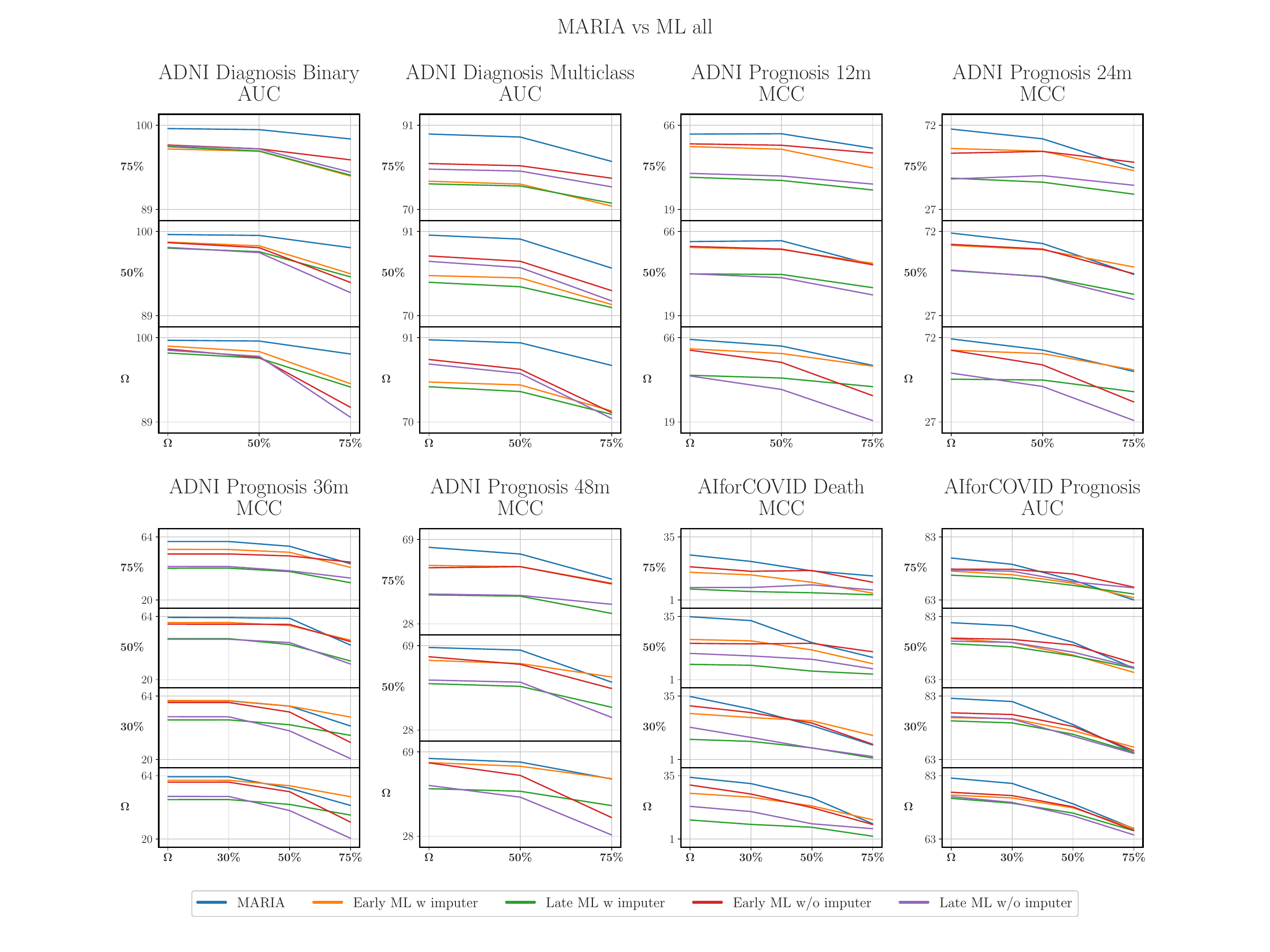}
    \caption{\modelName~vs. ML in the ``all missing'' scenario. Each plot, one for each task, reports different charts, one for each of the missing rates in the training set, showing how the performance (y-axis) changes as the missing rate in the test set increases.}
    \label{fig:MLall}
\end{figure*}

As an initial analysis, we compared \modelName~with traditional ML approaches, as depicted in \figurename s~\ref{fig:MLmod} and \ref{fig:MLall}.
These figures present the average performance of the models under investigation, categorized by their use of either an imputer or the MIA strategy (without an imputer) and further grouped into early and late fusion approaches.
Notably, \modelName~consistently demonstrates superior average performance across all levels of missing data in both the training and testing phases, maintaining its advantage even under ideal conditions where no additional missing data is introduced.

This consistent superiority highlights not only \modelName's distinct advantages over traditional ML models but also underscores the largely unexplored potential of DL methods in addressing incomplete data.
Moreover, the performance gap between \modelName~and its competitors widens as the missing data rate during training increases, whether in the ``missing modalities'' or ``all missing'' configurations.
This trend suggests that \modelName~is particularly resilient to varying missing data scenarios.
We attribute the improved performance of \modelName~to its robust regularization techniques, which enable the model to effectively learn how to handle diverse missing rates during training.

Additionally, an analysis of \figurename~\ref{fig:MLall} reveals that early fusion approaches generally outperform late fusion ones. 
This finding underscores the limitations of late fusion in capturing intercorrelations between features from different modalities.
Indeed, while late fusion can be compared to an ensemble of many experts, each specialized in a single modality and making independent predictions without communicating with each other, early fusion has a comprehensive view of the patient's information, similar to how a physician integrates all available data about a patient, allowing it to achieve superior performance as a result.

\subsection{\modelName~vs. DL}

\begin{figure*}[!t]
    \centering
    \includegraphics[width=\linewidth, trim={4cm 1cm 4cm 2cm}, clip]{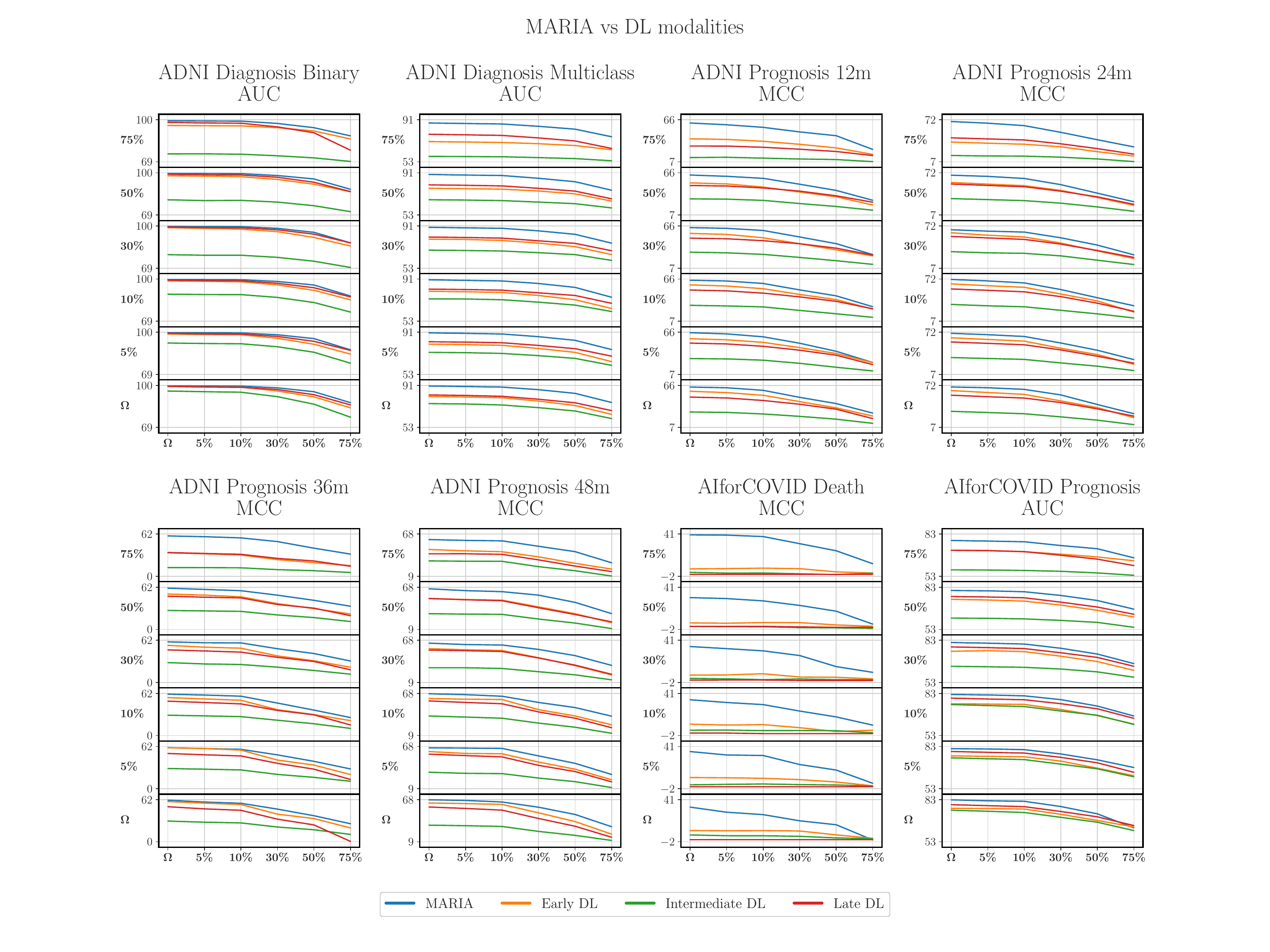}
    \caption{\modelName~vs. DL in the ``missing modalities'' scenario. Each plot, one for each task, reports different charts, one for each of the missing rates in the training set, showing how the performance (y-axis) changes as the missing rate in the test set increases.}
    \label{fig:DLmod}
\end{figure*}

\begin{figure*}[!t]
    \centering
    \includegraphics[width=\linewidth, trim={4cm 1cm 4cm 2cm}, clip]{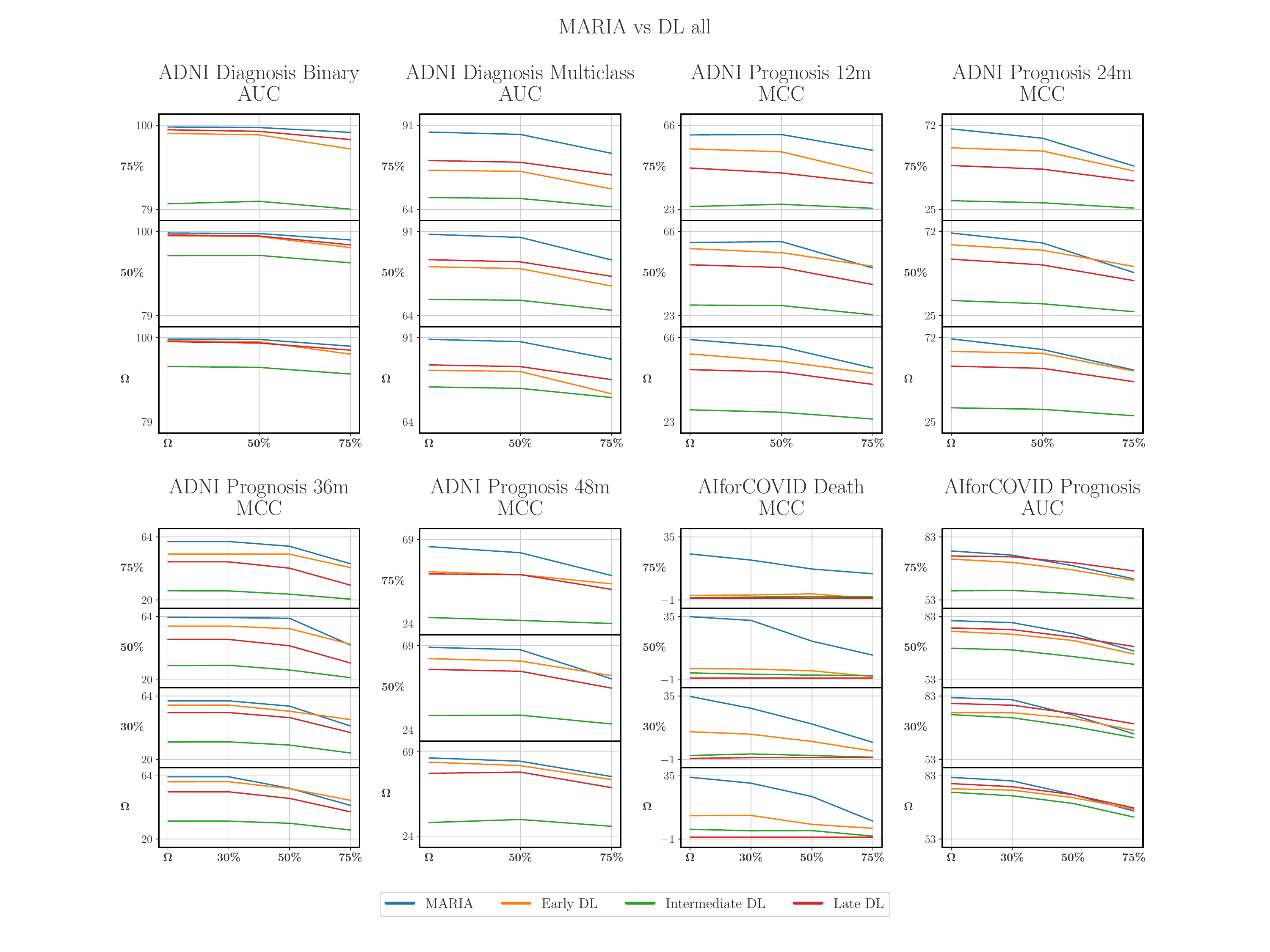}
    \caption{\modelName~vs. DL in the ``all missing'' scenario. Each plot, one for each task, reports different charts, one for each of the missing rates in the training set, showing how the performance (y-axis) changes as the missing rate in the test set increases.}
    \label{fig:DLall}
\end{figure*}

We also compared \modelName~to leading competitors from the DL domain, specifically designed to analyze tabular data.
In \figurename s~\ref{fig:DLmod} and \ref{fig:DLall}, we reported the average performance of these models when paired with an imputer and grouped into early and late fusion approaches.
Additionally, since no existing multimodal approaches are tailored for tabular data, we compared our method to intermediate fusion versions based on these models, as described in section~\ref{sec:competitors}.
Once again, \modelName~demonstrated superior performance across all levels of missing data in both experimental configurations.
Similar to the observations made in comparisons with ML-based approaches, the performance gap between \modelName~and its competitors widened as the missing rate in the training set increased.
However, unlike previous experiments, early and late fusion approaches exhibited comparable performance, indicating that both methodologies possess similar robustness to missing data.
This outcome highlights the capacity of DL techniques to derive meaningful and informative representations from data, even in incomplete scenarios.

In contrast, the intermediate fusion approaches struggled to match the performance of other methods, consistently failing to perform as well as early fusion models.
This suggests that intermediate fusion learning is unlikely to outperform its corresponding early and late fusion configurations when all modalities are tabular, particularly in scenarios with high missing rates.

\subsection{\modelName~vs. NAIM}

\begin{figure*}[!t]
    \centering
    \includegraphics[width=\linewidth, trim={4cm 1cm 4cm 2cm}, clip]{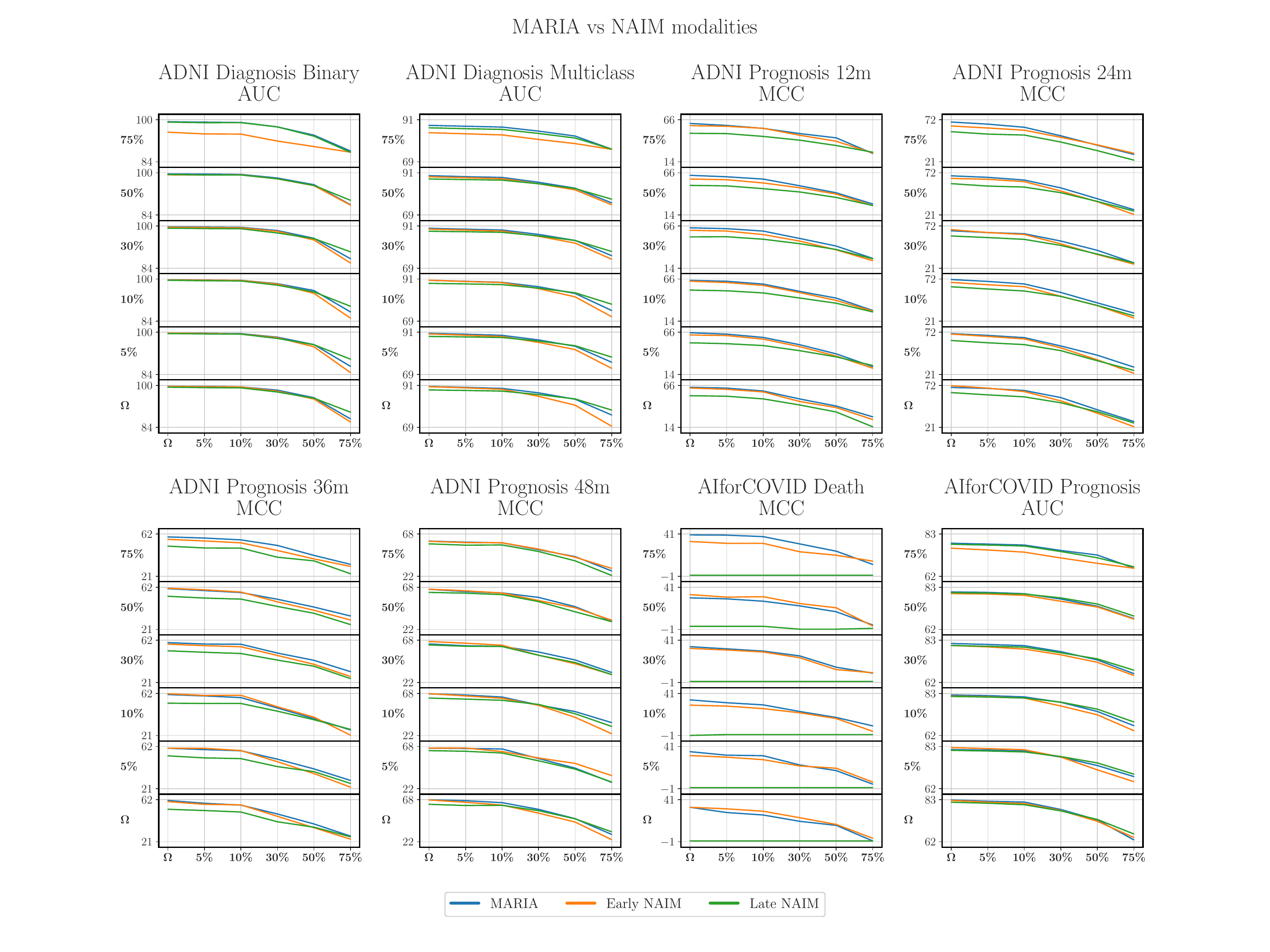}
    \caption{\modelName~vs. NAIM in the ``missing modalities'' scenario. Each plot, one for each task, reports different charts, one for each of the missing rates in the training set, showing how the performance (y-axis) changes as the missing rate in the test set increases.}
    \label{fig:NAIMmod}
\end{figure*}

\begin{figure*}[!t]
    \centering
    \includegraphics[width=\linewidth, trim={4cm 1cm 4cm 2cm}, clip]{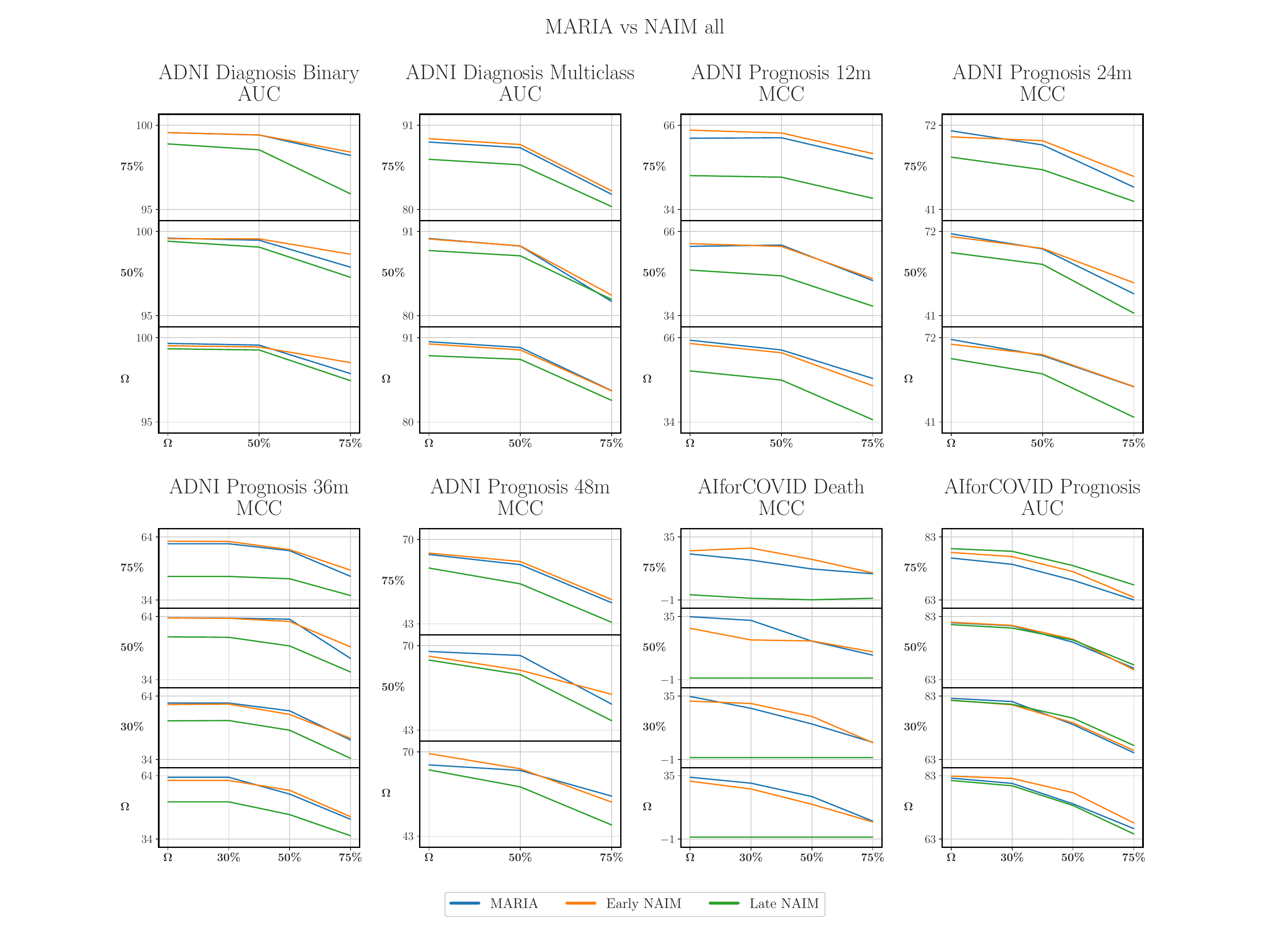}
    \caption{\modelName~vs. NAIM in the ``all missing'' scenario. Each plot, one for each task, reports different charts, one for each of the missing rates in the training set, showing how the performance (y-axis) changes as the missing rate in the test set increases.}
    \label{fig:NAIMall}
\end{figure*}

As a final analysis, we compared the proposed intermediate fusion approach with its respective early and late fusion counterparts, all based on the NAIM model~\cite{bib:NAIM}.
As in the previous analyses, we used the same evaluation framework to compare the different approaches (\figurename s~\ref{fig:NAIMmod} and \ref{fig:NAIMall}).

Interestingly, and as previously noted but now more pronounced, the intermediate fusion approach struggles to consistently outperform the early fusion approach, particularly in the context of tabular data.
This suggests that early fusion may offer specific advantages when handling highly structured tabular datasets, where the model benefits from a unified representation of all input features from the beginning.
In such contexts, the expected benefits of intermediate fusion appear less significant, potentially due to the inherent heterogeneity of tabular features, which may not require the additional representational flexibility provided by intermediate fusion.

By contrast, the late fusion approach, especially in the "all missing" configuration, rarely achieves performance comparable to the other two approaches.
This limitation highlights the challenges faced by late fusion in capturing intercorrelations between features from different modalities, particularly when data completeness is severely compromised.

Overall, these results underscore the trade-offs between fusion strategies, suggesting that the choice between early, intermediate, and late fusion should be guided by the specific characteristics of the data modalities.
For structured tabular data, early fusion seems to provide an optimal balance between simplicity and performance.
By contrast, intermediate and late fusion approaches may be more beneficial in scenarios involving heterogeneous or unstructured data sources.

In \ref{app1}, we present detailed tables for both types of experiments, showing the average performance in terms of AUC and MCC computed across the respective tasks under consideration.
The tables include results at different percentages of missing values (reported in the columns).
The first rows indicate the percentage of missing values used during training and testing, while the subsequent rows report the performance for each combination of fusion strategy, model, and missing data handling technique.
To facilitate readability, we highlighted in bold the best performance for each combination of missing rates.

As presented in \tablename~\ref{tab:MOD_res}, which summarizes the results for the ``missing modalities'' setting, the first table shows the average AUC scores for the ADNI diagnosis and AIforCOVID prognosis tasks. 
In this context, \modelName~outperforms its competitors in $69.4\%$ of cases (25 out of 36). 
Similarly, the second table reports the average MCC scores for the ADNI prognosis and AIforCOVID death tasks, where \modelName~achieves superior performance in $61.1\%$ of cases (22 out of 36).

In contrast, \tablename~\ref{tab:ALL_res}, which details the results for the ``all missing'' setting, indicates that \modelName~exhibits stronger performance in fewer instances: $37.5\%$ (6 out of 16) for tasks evaluated using AUC (first table) and $12.5\%$ (2 out of 16) for tasks assessed with MCC (second table). 
This difference is likely due to the "all missing" setting generally involving less severe information loss compared to the "missing modalities" setting. 
These findings suggest that the advantages of \modelName~are particularly pronounced in scenarios characterized by significant modality loss.

These analyses demonstrate the robustness of the \modelName~model, particularly under conditions with high missing rates, a common challenge in medical applications where patient data are often incomplete or inconsistently available.
By effectively leveraging only the available information, the \modelName~model enhances diagnostic accuracy and decision support in healthcare, ultimately leading to improved patient outcomes.
Across a wide range of experimental configurations and degrees of missingness, the model consistently outperforms competing approaches, including both traditional ML and DL models.


\section{Conclusions and Discussion}\label{sec:conclusion}

In this study, we introduced \modelName, a novel transformer-based model designed to tackle the challenges of incomplete multimodal data, especially in healthcare. 
\modelName~employs an intermediate fusion strategy, integrating data from multiple incomplete modalities through a masked self-attention mechanism that selectively focuses on available information while ignoring missing parts. 
This approach not only avoids the disadvantages of synthetic imputation but also ensures a robust predictive performance even in the presence of severe data missingness. 
The results show that \modelName~is effective across multiple diagnostic and prognostic tasks, consistently outperforming traditional ML and DL models by adapting to various missing data scenarios during both training and inference.

Despite these promising results, some limitations need to be addressed. 
One of the key challenges of \modelName~lies in its computational complexity. 
Although the model eliminates the need for imputers, the use of masked attention mechanisms and the intermediate fusion strategy requires substantial computational resources, which may limit its scalability in low-resource environments or when applied to extremely large datasets. 
To mitigate this, future work will focus on enhancing \modelName's scalability and efficiency, potentially by incorporating more lightweight attention mechanisms or by developing hybrid techniques that blend intermediate fusion with other fusion strategies in a computationally feasible manner.

Furthermore, while \modelName's design is effective for tabular data, its generalization to other types of multimodal inputs, such as imaging or textual data, has not been fully explored.
To address this, future research will aim to expand \modelName's applicability to other forms of multimodal data, including medical images and clinical notes, which would ensure broader usability across diverse healthcare datasets.


\section*{Acknowledgment}
Camillo Maria Caruso is a Ph.D. student enrolled in the National Ph.D. in Artificial Intelligence, XXXVII cycle, course on Health and life sciences, organized by Università Campus Bio-Medico di Roma.

This work was partially founded by: 
i) Università Campus Bio-Medico di Roma under the program ``University Strategic Projects'' within the project ``AI-powered Digital Twin for next-generation lung cancEr cAre (IDEA)''; 
ii) from PRIN 2022 MUR 20228MZFAA-AIDA (CUP C53D23003620008); 
iii) from PRIN PNRR 2022 MUR P2022P3CXJ-PICTURE (CUP C53D23009280001);
iv) from PNRR MUR project PE0000013-FAIR;
v) Project ECS 0000024 Rome Technopole (CUP C83C22000510001).

Resources are provided by the National Academic Infrastructure for Supercomputing in Sweden (NAISS) and the Swedish National Infrastructure for Computing (SNIC) at Alvis @ C3SE, partially funded by the Swedish Research Council through grant agreements no. 2022-06725 and no. 2018-05973.

Data collection and sharing for this project was funded by the Alzheimer's Disease Neuroimaging Initiative (ADNI) (National Institutes of Health Grant U01 AG024904) and DOD ADNI (Department of Defense award number W81XWH-12-2-0012). ADNI is funded by the National Institute on Aging, the National Institute of Biomedical Imaging and Bioengineering, and through generous contributions from the following: AbbVie, Alzheimer’s Association; Alzheimer’s Drug Discovery Foundation; Araclon Biotech; BioClinica, Inc.; Biogen; Bristol-Myers Squibb Company; CereSpir, Inc.; Cogstate; Eisai Inc.; Elan Pharmaceuticals, Inc.; Eli Lilly and Company; EuroImmun; F. Hoffmann-La Roche Ltd and its affiliated company Genentech, Inc.; Fujirebio; GE Healthcare; IXICO Ltd.; Janssen Alzheimer Immunotherapy Research \& Development, LLC.; Johnson \& Johnson Pharmaceutical Research \& Development LLC.; Lumosity; Lundbeck; Merck \& Co., Inc.; Meso Scale Diagnostics, LLC.; NeuroRx Research; Neurotrack Technologies; Novartis Pharmaceuticals Corporation; Pfizer Inc.; Piramal Imaging; Servier; Takeda Pharmaceutical Company; and Transition Therapeutics. The Canadian Institutes of Health Research is providing funds to support ADNI clinical sites in Canada. Private sector contributions are facilitated by the Foundation for the National Institutes of Health (\url{www.fnih.org}). The grantee organization is the Northern California Institute for Research and Education, and the study is coordinated by the Alzheimer’s Therapeutic Research Institute at the University of Southern California. ADNI data are disseminated by the Laboratory for Neuro Imaging at the University of Southern California.

\section*{Author Contributions}
\textbf{Camillo Maria Caruso:} Conceptualization, Data curation, Formal analysis, Investigation, Methodology, Software, Validation, Visualization, Writing – original draft, Writing – review \& editing;
\textbf{Paolo Soda:} Conceptualization, Funding acquisition, Investigation, Methodology, Project administration, Resources, Supervision, Writing – review \& editing;
\textbf{Valerio Guarrasi:} Conceptualization, Formal analysis, Funding acquisition, Investigation, Methodology, Project administration, Resources, Supervision, Validation, Visualization, Writing – original draft, Writing – review \& editing.
	
\bibliographystyle{elsarticle-num} 
\bibliography{bibliography}


\appendix
\clearpage
\onecolumn
\section{Additional results}\label{app1}
\vfill
\begin{figure}[!htbp]
\centering
\begin{adjustbox}{angle=90}
\begin{minipage}{.7\textheight}
\centering
\resizebox{\linewidth}{!}{
\begin{tabular}{lll|cccccc|cccccc|cccccc|cccccc|cccccc|cccccc}
\toprule
 \textbf{AUC} &  &  & \multicolumn{6}{c|}{\textbf{Train missing percentage:} $\Omega$} & \multicolumn{6}{c|}{\textbf{Train missing percentage:} 5\%} & \multicolumn{6}{c|}{\textbf{Train missing percentage:} 10\%} & \multicolumn{6}{c|}{\textbf{Train missing percentage:} 30\%} & \multicolumn{6}{c|}{\textbf{Train missing percentage:} 50\%} & \multicolumn{6}{c}{\textbf{Train missing percentage:} 75\%} \\\hline
 &  &  & \multicolumn{6}{c|}{\textbf{Test missing percentage:}} & \multicolumn{6}{c|}{\textbf{Test missing percentage:}} & \multicolumn{6}{c|}{\textbf{Test missing percentage:}} & \multicolumn{6}{c|}{\textbf{Test missing percentage:}} & \multicolumn{6}{c|}{\textbf{Test missing percentage:}} & \multicolumn{6}{c}{\textbf{Test missing percentage:}} \\
\textbf{Fusion Strategy} & \textbf{Model} & \textbf{Imputer} & $\Omega$ & 5\% & 10\% & 30\% & 50\% & 75\% & $\Omega$ & 5\% & 10\% & 30\% & 50\% & 75\% & $\Omega$ & 5\% & 10\% & 30\% & 50\% & 75\% & $\Omega$ & 5\% & 10\% & 30\% & 50\% & 75\% & $\Omega$ & 5\% & 10\% & 30\% & 50\% & 75\% & $\Omega$ & 5\% & 10\% & 30\% & 50\% & 75\% \\
\midrule
\multirow[c]{14}{*}{\textbf{Early}} & AdaBoost & with & 83.89 & 83.53 & 83.10 & 80.24 & 76.17 & 69.71 & 84.27 & 83.89 & 83.43 & 80.27 & 76.46 & 70.63 & 84.12 & 83.44 & 83.27 & 79.91 & 76.56 & 69.73 & 81.96 & 81.67 & 80.98 & 78.33 & 75.11 & 69.47 & 81.13 & 80.66 & 80.38 & 78.17 & 75.52 & 70.25 & 80.44 & 79.98 & 79.67 & 77.65 & 74.74 & 71.19 \\
\cline{2-39}
 & \multirow[c]{2}{*}{DecisionTree} & with & 76.40 & 75.59 & 74.91 & 70.88 & 66.96 & 60.07 & 74.04 & 73.48 & 73.03 & 70.33 & 67.36 & 61.25 & 74.41 & 73.90 & 73.40 & 70.43 & 66.69 & 60.79 & 71.53 & 71.20 & 70.70 & 68.35 & 66.63 & 61.50 & 70.01 & 69.74 & 69.23 & 67.54 & 66.37 & 61.87 & 73.05 & 72.25 & 71.68 & 68.97 & 65.89 & 62.79 \\
 &  & without & 78.33 & 77.22 & 76.51 & 71.92 & 67.45 & 60.68 & 78.12 & 77.23 & 76.18 & 71.71 & 67.15 & 60.62 & 78.08 & 77.47 & 77.24 & 73.24 & 68.06 & 60.84 & 76.35 & 76.23 & 75.86 & 72.71 & 69.79 & 64.83 & 76.19 & 75.75 & 75.29 & 73.11 & 71.55 & 66.33 & 76.32 & 75.78 & 75.57 & 74.25 & 71.72 & 68.38 \\
\cline{2-39}
 & FTTransformer & with & 87.47 & 87.15 & 86.65 & 83.60 & 79.15 & 72.01 & 86.50 & 86.27 & 85.86 & 83.11 & 78.54 & 71.60 & 86.54 & 86.26 & 85.92 & 83.28 & 78.96 & 71.88 & 86.42 & 86.30 & 85.43 & 82.89 & 79.19 & 72.39 & 85.49 & 85.16 & 84.44 & 82.07 & 78.67 & 72.95 & 71.80 & 71.56 & 71.19 & 69.94 & 68.12 & 65.41 \\
\cline{2-39}
 & \multirow[c]{2}{*}{HistGradientBoost} & with & 88.05 & 87.39 & 86.80 & 83.60 & 79.25 & 71.36 & 87.85 & 87.47 & 86.87 & 83.31 & 78.47 & 71.18 & 88.00 & 87.33 & 86.64 & 82.95 & 78.76 & 70.89 & 86.78 & 86.49 & 85.83 & 82.90 & 79.30 & 70.62 & 84.91 & 84.46 & 83.98 & 81.67 & 78.24 & 70.82 & 79.48 & 79.24 & 78.78 & 76.45 & 74.34 & 70.42 \\
 &  & without & 89.99 & 89.22 & 88.44 & 84.35 & 79.59 & 71.70 & 89.89 & 89.33 & 88.70 & 85.17 & 80.14 & 72.19 & 90.00 & 89.44 & 88.90 & 85.31 & 80.25 & 71.13 & 89.00 & 88.38 & 87.64 & 84.38 & 80.50 & 72.31 & 87.26 & 86.74 & 86.19 & 82.90 & 79.15 & 71.80 & 81.88 & 81.37 & 80.94 & 78.63 & 76.12 & 71.71 \\
\cline{2-39}
 & MLP & with & 84.98 & 84.62 & 84.14 & 80.82 & 76.63 & 69.86 & 84.96 & 84.44 & 84.04 & 80.81 & 76.89 & 69.40 & 84.22 & 83.93 & 83.48 & 80.58 & 76.85 & 69.50 & 83.88 & 83.63 & 83.15 & 80.45 & 76.77 & 69.87 & 82.89 & 82.27 & 81.74 & 79.53 & 76.53 & 70.47 & 86.27 & 85.82 & 85.33 & 83.33 & 80.29 & 74.55 \\
\cline{2-39}
 & \multirow[c]{2}{*}{RandomForest} & with & 88.75 & 88.17 & 87.66 & 84.08 & 79.67 & 71.68 & 88.81 & 88.28 & 87.71 & 83.39 & 78.69 & 71.48 & 88.49 & 88.15 & 87.79 & 83.82 & 79.73 & 71.34 & 87.78 & 87.62 & 86.87 & 83.60 & 79.44 & 71.04 & 87.05 & 86.62 & 85.91 & 82.41 & 78.59 & 71.63 & 87.13 & 86.37 & 85.60 & 82.38 & 78.28 & 71.74 \\
 &  & without & 90.78 & 90.26 & 89.69 & 86.32 & 81.63 & 73.58 & \textbf{90.65} & 90.14 & 89.49 & 86.11 & 81.42 & 72.98 & 90.68 & 90.23 & 89.75 & 86.48 & 81.77 & 72.85 & \textbf{90.39} & 89.93 & 89.45 & 86.45 & 82.46 & 74.17 & \textbf{90.03} & \textbf{89.75} & \textbf{89.30} & 86.63 & 83.11 & 75.48 & 88.25 & 87.84 & 87.40 & 85.05 & 81.82 & 76.36 \\
\cline{2-39}
 & SVM & with & 86.43 & 86.08 & 85.70 & 82.76 & 78.28 & 71.80 & 86.56 & 86.17 & 86.01 & 82.61 & 78.37 & 71.85 & 86.71 & 86.39 & 86.09 & 82.99 & 78.87 & 71.97 & 86.74 & 86.57 & 86.04 & 83.20 & 79.21 & 72.07 & 86.15 & 85.70 & 85.21 & 83.01 & 78.90 & 72.87 & 84.82 & 84.42 & 83.90 & 81.73 & 78.42 & 74.01 \\
\cline{2-39}
 & TABTransformer & with & 84.65 & 84.32 & 83.92 & 80.98 & 77.43 & 70.57 & 84.06 & 83.72 & 83.58 & 80.98 & 76.96 & 70.34 & 83.91 & 83.56 & 83.19 & 80.50 & 76.26 & 70.09 & 83.16 & 82.94 & 82.39 & 80.23 & 76.91 & 70.77 & 82.60 & 82.18 & 81.45 & 79.63 & 76.38 & 70.82 & 86.35 & 85.94 & 85.39 & 83.34 & 80.35 & 74.80 \\
\cline{2-39}
 & TabNet & with & 85.62 & 85.11 & 84.90 & 81.50 & 76.70 & 69.97 & 84.76 & 84.21 & 83.81 & 81.07 & 76.37 & 69.40 & 84.41 & 84.10 & 83.52 & 80.32 & 76.38 & 69.00 & 83.83 & 84.01 & 83.03 & 80.49 & 76.15 & 69.65 & 81.42 & 81.28 & 81.01 & 78.85 & 75.33 & 70.27 & 73.50 & 73.77 & 73.22 & 73.03 & 72.80 & 70.24 \\
\cline{2-39}
 & \multirow[c]{2}{*}{XGBoost} & with & 88.00 & 87.32 & 86.71 & 82.68 & 77.94 & 70.04 & 88.07 & 87.56 & 87.07 & 83.36 & 78.33 & 70.57 & 87.80 & 87.19 & 86.77 & 83.08 & 78.53 & 70.90 & 86.44 & 86.16 & 85.60 & 82.86 & 79.19 & 71.17 & 85.16 & 84.86 & 84.35 & 81.91 & 79.33 & 72.36 & 85.87 & 85.47 & 84.99 & 82.65 & 79.65 & 74.40 \\
 &  & without & 90.07 & 89.41 & 88.46 & 84.03 & 78.75 & 70.43 & 90.10 & 89.64 & 89.07 & 86.22 & 82.23 & 74.02 & 89.71 & 89.39 & 89.02 & 86.27 & 82.63 & 74.28 & 88.97 & 88.68 & 88.39 & 86.10 & 83.46 & 76.07 & 89.05 & 88.81 & 88.42 & 86.02 & 83.26 & 76.98 & 88.15 & 87.66 & 87.36 & 85.23 & 82.28 & \textbf{76.95} \\
\cline{1-39} \cline{2-39}
\multirow[c]{5}{*}{\textbf{Intermediate}} & FTTransformer & with & 80.43 & 79.98 & 79.25 & 75.57 & 71.45 & 63.68 & 69.22 & 68.99 & 68.54 & 66.57 & 64.75 & 60.19 & 67.73 & 67.60 & 67.16 & 65.41 & 63.66 & 59.58 & 57.28 & 57.12 & 56.86 & 55.90 & 55.17 & 53.25 & 52.86 & 52.83 & 52.68 & 52.12 & 51.95 & 50.85 & 50.00 & 50.00 & 50.00 & 50.00 & 50.00 & 50.00 \\
\cline{2-39}
 & MLP & with & 86.62 & 86.16 & 85.60 & 82.97 & 79.35 & 71.43 & 87.19 & 86.80 & 86.28 & 83.83 & 80.43 & 72.24 & 86.65 & 86.15 & 85.59 & 83.19 & 79.95 & 72.48 & 78.66 & 78.52 & 78.36 & 77.12 & 74.97 & 69.38 & 76.84 & 76.64 & 76.29 & 75.25 & 73.66 & 69.19 & 67.90 & 67.75 & 67.54 & 66.79 & 65.43 & 62.12 \\
\cline{2-39}
 & \modelName & without & \textbf{90.94} & \textbf{90.55} & \textbf{90.20} & \textbf{87.78} & \textbf{83.97} & 75.24 & 90.49 & \textbf{90.26} & \textbf{89.86} & \textbf{87.59} & \textbf{84.17} & 76.80 & \textbf{90.76} & \textbf{90.45} & \textbf{90.03} & \textbf{87.94} & \textbf{84.38} & 76.36 & 90.28 & \textbf{89.94} & \textbf{89.56} & \textbf{87.39} & \textbf{84.00} & 76.49 & 89.92 & 89.63 & 89.25 & \textbf{87.07} & \textbf{84.00} & 76.86 & \textbf{88.57} & \textbf{88.22} & \textbf{87.88} & \textbf{85.72} & \textbf{83.05} & 76.61 \\
\cline{2-39}
 & TABTransformer & with & 85.20 & 84.41 & 83.69 & 80.66 & 76.25 & 68.57 & 84.17 & 83.55 & 82.98 & 80.24 & 76.08 & 68.24 & 84.95 & 84.37 & 83.91 & 80.87 & 76.70 & 68.84 & 83.25 & 82.86 & 82.33 & 79.97 & 76.71 & 69.11 & 82.73 & 82.33 & 81.89 & 79.62 & 76.23 & 70.78 & 81.35 & 81.02 & 80.52 & 78.40 & 75.24 & 69.84 \\
\cline{2-39}
 & TabNet & with & 75.54 & 75.07 & 74.20 & 71.40 & 67.48 & 61.05 & 79.15 & 78.69 & 78.30 & 75.51 & 71.77 & 65.70 & 76.87 & 76.54 & 76.15 & 73.40 & 70.03 & 63.66 & 64.28 & 63.48 & 63.32 & 62.18 & 59.52 & 56.47 & 64.81 & 64.04 & 64.01 & 63.00 & 60.77 & 56.24 & 54.49 & 54.74 & 54.40 & 53.92 & 53.61 & 53.97 \\
\cline{1-39} \cline{2-39}
\multirow[c]{14}{*}{\textbf{Late}} & AdaBoost & with & 83.94 & 82.90 & 82.19 & 78.55 & 75.09 & 69.87 & 83.77 & 83.15 & 82.85 & 80.83 & 78.48 & 73.30 & 83.01 & 82.46 & 82.05 & 80.24 & 77.74 & 73.06 & 82.47 & 82.00 & 81.72 & 79.54 & 77.22 & 72.41 & 82.23 & 81.81 & 81.50 & 79.46 & 76.74 & 71.12 & 83.08 & 82.48 & 81.90 & 79.63 & 76.62 & 71.13 \\
\cline{2-39}
 & \multirow[c]{2}{*}{DecisionTree} & with & 77.36 & 76.58 & 75.74 & 72.42 & 69.04 & 62.32 & 77.64 & 77.21 & 76.53 & 74.12 & 71.32 & 63.76 & 77.30 & 76.98 & 76.61 & 74.37 & 71.36 & 63.97 & 76.51 & 76.21 & 76.11 & 74.01 & 71.07 & 64.48 & 76.38 & 76.32 & 76.02 & 72.94 & 70.08 & 62.83 & 75.52 & 74.92 & 74.27 & 71.53 & 69.12 & 63.08 \\
 &  & without & 83.18 & 82.38 & 81.92 & 78.24 & 75.01 & 68.07 & 83.25 & 82.77 & 82.26 & 79.26 & 76.18 & 69.68 & 83.07 & 82.66 & 82.17 & 79.60 & 76.86 & 71.26 & 82.12 & 81.76 & 81.53 & 79.19 & 75.69 & 69.46 & 81.90 & 81.60 & 81.47 & 78.84 & 75.99 & 69.75 & 80.23 & 79.78 & 79.30 & 76.76 & 73.75 & 67.96 \\
\cline{2-39}
 & FTTransformer & with & 87.15 & 86.67 & 86.07 & 83.56 & 80.34 & 73.76 & 88.13 & 87.69 & 87.25 & 85.02 & 81.73 & 75.51 & 87.92 & 87.44 & 87.03 & 84.87 & 81.86 & 75.26 & 86.76 & 86.34 & 85.85 & 83.66 & 80.78 & 74.13 & 85.08 & 84.73 & 84.27 & 82.00 & 79.00 & 72.82 & 82.61 & 82.18 & 81.60 & 79.10 & 76.12 & 67.04 \\
\cline{2-39}
 & \multirow[c]{2}{*}{HistGradientBoost} & with & 83.29 & 82.42 & 81.63 & 77.80 & 74.38 & 68.17 & 82.70 & 82.11 & 81.33 & 78.64 & 75.33 & 68.73 & 82.82 & 82.21 & 81.52 & 78.96 & 75.88 & 69.45 & 82.78 & 82.18 & 81.61 & 79.07 & 75.81 & 70.19 & 81.57 & 81.15 & 80.75 & 78.24 & 75.30 & 68.93 & 80.72 & 80.17 & 79.61 & 77.19 & 74.04 & 68.01 \\
 &  & without & 84.66 & 83.90 & 83.14 & 79.65 & 75.97 & 69.85 & 84.51 & 84.04 & 83.44 & 80.67 & 77.54 & 70.45 & 84.43 & 83.95 & 83.37 & 80.66 & 77.72 & 71.19 & 84.42 & 83.95 & 83.58 & 80.71 & 77.50 & 71.75 & 83.75 & 83.26 & 82.91 & 80.25 & 77.32 & 70.55 & 81.79 & 81.08 & 80.52 & 77.85 & 74.60 & 69.03 \\
\cline{2-39}
 & MLP & with & 87.13 & 86.66 & 86.13 & 83.55 & 80.03 & 72.68 & 87.52 & 87.13 & 86.71 & 84.26 & 80.89 & 74.23 & 87.26 & 86.85 & 86.49 & 84.18 & 81.17 & 74.39 & 87.16 & 86.82 & 86.47 & 84.18 & 81.35 & 74.96 & 86.49 & 86.09 & 85.64 & 83.31 & 79.99 & 73.95 & 83.93 & 83.49 & 82.99 & 80.59 & 77.01 & 71.81 \\
\cline{2-39}
 & \multirow[c]{2}{*}{RandomForest} & with & 84.92 & 84.26 & 83.34 & 79.77 & 75.92 & 68.86 & 84.63 & 84.29 & 83.81 & 81.66 & 78.85 & 72.49 & 84.48 & 84.20 & 83.76 & 81.88 & 79.08 & 72.86 & 84.15 & 83.88 & 83.49 & 81.52 & 78.84 & 73.27 & 83.49 & 83.33 & 82.94 & 80.93 & 78.18 & 72.38 & 83.18 & 82.68 & 82.25 & 80.05 & 77.30 & 71.50 \\
 &  & without & 88.70 & 88.21 & 87.70 & 84.84 & 81.37 & 75.23 & 89.00 & 88.66 & 88.29 & 86.22 & 83.39 & \textbf{77.60} & 88.89 & 88.56 & 88.18 & 86.10 & 83.43 & \textbf{77.89} & 88.70 & 88.38 & 88.11 & 85.99 & 83.22 & \textbf{77.89} & 88.36 & 88.04 & 87.81 & 85.85 & 83.20 & \textbf{77.16} & 87.93 & 87.53 & 87.14 & 84.96 & 82.05 & 76.53 \\
\cline{2-39}
 & SVM & with & 88.13 & 87.67 & 87.14 & 84.47 & 80.92 & 73.94 & 88.21 & 87.71 & 87.22 & 84.73 & 81.00 & 73.78 & 88.12 & 87.60 & 87.13 & 84.52 & 80.68 & 72.95 & 88.08 & 87.55 & 87.04 & 84.36 & 80.29 & 72.88 & 87.42 & 86.84 & 86.27 & 83.31 & 78.96 & 71.46 & 85.28 & 84.15 & 83.23 & 79.04 & 73.58 & 66.62 \\
\cline{2-39}
 & TABTransformer & with & 86.42 & 85.77 & 85.18 & 82.57 & 78.39 & 71.01 & 85.28 & 84.74 & 84.26 & 81.68 & 77.87 & 70.41 & 85.64 & 85.13 & 84.62 & 81.99 & 78.59 & 70.96 & 84.96 & 84.34 & 83.81 & 81.33 & 78.24 & 71.47 & 84.04 & 83.61 & 83.19 & 80.61 & 77.21 & 70.76 & 82.76 & 82.36 & 81.91 & 79.37 & 75.60 & 63.81 \\
\cline{2-39}
 & TabNet & with & 87.14 & 86.71 & 86.10 & 83.56 & 80.10 & 73.73 & 87.19 & 86.73 & 86.37 & 84.01 & 80.60 & 74.09 & 86.90 & 86.52 & 86.14 & 83.87 & 80.87 & 74.48 & 85.72 & 85.37 & 84.90 & 82.70 & 79.79 & 73.70 & 85.03 & 84.73 & 84.32 & 82.25 & 79.10 & 72.43 & 80.57 & 80.20 & 79.82 & 77.21 & 74.48 & 68.31 \\
\cline{2-39}
 & \multirow[c]{2}{*}{XGBoost} & with & 87.65 & 87.03 & 86.43 & 83.26 & 79.46 & 72.08 & 87.36 & 86.91 & 86.41 & 84.08 & 81.17 & 74.80 & 87.19 & 86.75 & 86.42 & 84.05 & 80.90 & 74.87 & 86.60 & 86.29 & 85.88 & 83.62 & 80.61 & 74.52 & 86.20 & 85.93 & 85.52 & 83.10 & 80.17 & 74.00 & 85.34 & 84.97 & 84.33 & 81.89 & 78.91 & 72.71 \\
 &  & without & 89.05 & 88.46 & 88.05 & 85.28 & 81.87 & \textbf{75.66} & 88.99 & 88.61 & 88.21 & 85.97 & 82.97 & 77.14 & 88.89 & 88.57 & 88.22 & 86.02 & 83.00 & 77.35 & 88.46 & 88.17 & 87.83 & 85.54 & 82.39 & 76.91 & 88.23 & 87.92 & 87.64 & 85.27 & 82.27 & 76.12 & 86.25 & 85.91 & 85.41 & 82.99 & 79.82 & 74.26 \\
\bottomrule
\end{tabular}

}

\vspace*{.5cm}

\resizebox{\linewidth}{!}{
\begin{tabular}{lll|cccccc|cccccc|cccccc|cccccc|cccccc|cccccc}
\toprule
 \textbf{MCC} &  &  & \multicolumn{6}{c|}{\textbf{Train missing percentage:} $\Omega$} & \multicolumn{6}{c|}{\textbf{Train missing percentage:} 5\%} & \multicolumn{6}{c|}{\textbf{Train missing percentage:} 10\%} & \multicolumn{6}{c|}{\textbf{Train missing percentage:} 30\%} & \multicolumn{6}{c|}{\textbf{Train missing percentage:} 50\%} & \multicolumn{6}{c}{\textbf{Train missing percentage:} 75\%} \\\hline
 &  &  & \multicolumn{6}{c|}{\textbf{Test missing percentage:}} & \multicolumn{6}{c|}{\textbf{Test missing percentage:}} & \multicolumn{6}{c|}{\textbf{Test missing percentage:}} & \multicolumn{6}{c|}{\textbf{Test missing percentage:}} & \multicolumn{6}{c|}{\textbf{Test missing percentage:}} & \multicolumn{6}{c}{\textbf{Test missing percentage:}} \\
\textbf{Fusion Strategy} & \textbf{Model} & \textbf{Imputer} & $\Omega$ & 5\% & 10\% & 30\% & 50\% & 75\% & $\Omega$ & 5\% & 10\% & 30\% & 50\% & 75\% & $\Omega$ & 5\% & 10\% & 30\% & 50\% & 75\% & $\Omega$ & 5\% & 10\% & 30\% & 50\% & 75\% & $\Omega$ & 5\% & 10\% & 30\% & 50\% & 75\% & $\Omega$ & 5\% & 10\% & 30\% & 50\% & 75\% \\
\midrule
\multirow[c]{14}{*}{\textbf{Early}} & AdaBoost & with & 54.79 & 52.72 & 50.17 & 40.24 & 33.64 & 20.94 & 52.81 & 51.37 & 49.36 & 38.99 & 31.18 & 22.48 & 53.05 & 50.56 & 48.38 & 39.52 & 32.32 & 20.22 & 48.93 & 46.92 & 44.84 & 36.28 & 28.66 & 20.09 & 43.57 & 42.27 & 40.46 & 33.99 & 27.48 & 18.45 & 28.35 & 28.29 & 29.33 & 25.76 & 22.07 & 15.53 \\
\cline{2-39}
 & \multirow[c]{2}{*}{DecisionTree} & with & 47.80 & 46.58 & 44.73 & 38.54 & 31.37 & 19.83 & 46.20 & 44.75 & 42.88 & 34.76 & 29.04 & 18.01 & 44.65 & 43.66 & 41.97 & 33.92 & 27.69 & 19.63 & 40.69 & 39.81 & 39.34 & 33.36 & 26.16 & 17.39 & 36.17 & 35.78 & 33.81 & 29.31 & 23.39 & 14.03 & 44.41 & 42.97 & 40.94 & 34.06 & 28.85 & 18.00 \\
 &  & without & 50.50 & 48.47 & 45.60 & 36.30 & 30.37 & 19.03 & 47.92 & 46.89 & 44.33 & 36.91 & 29.49 & 17.91 & 49.06 & 48.16 & 46.02 & 37.68 & 29.18 & 19.46 & 47.25 & 45.69 & 43.15 & 37.75 & 29.75 & 21.54 & 43.04 & 42.85 & 41.62 & 36.97 & 30.84 & 20.99 & 48.37 & 47.12 & 45.26 & 38.39 & 31.58 & 20.63 \\
\cline{2-39}
 & FTTransformer & with & 52.35 & 50.42 & 48.16 & 38.19 & 30.47 & 18.55 & 51.48 & 49.63 & 48.08 & 38.17 & 31.07 & 18.41 & 52.10 & 50.03 & 48.29 & 38.16 & 31.14 & 19.27 & 50.93 & 48.56 & 46.44 & 37.76 & 29.58 & 19.99 & 47.92 & 45.86 & 43.84 & 34.36 & 26.92 & 18.87 & 8.45 & 8.03 & 7.63 & 6.28 & 4.55 & 2.49 \\
\cline{2-39}
 & \multirow[c]{2}{*}{HistGradientBoost} & with & 58.61 & 56.60 & \textbf{54.52} & 44.52 & 34.87 & 22.48 & 58.19 & 56.04 & 54.43 & 44.89 & 36.42 & 22.96 & 57.67 & 55.52 & 53.37 & 42.55 & 34.75 & 23.08 & 55.49 & 53.72 & 51.84 & 42.97 & 32.90 & 22.60 & 54.55 & 52.47 & 51.01 & 42.03 & 34.13 & 22.60 & 27.61 & 27.33 & 27.29 & 25.40 & 22.49 & 22.41 \\
 &  & without & 57.42 & 55.17 & 51.54 & 42.20 & 33.53 & \textbf{22.91} & 58.58 & 57.29 & 54.86 & 46.57 & 35.53 & 22.72 & 58.26 & 56.28 & 54.31 & 47.40 & 37.13 & 23.32 & 55.16 & 53.87 & 52.76 & 45.62 & 36.21 & 24.98 & 55.90 & 54.47 & 51.92 & 43.48 & 34.62 & 23.17 & 26.38 & 25.68 & 25.19 & 24.03 & 21.81 & 24.69 \\
\cline{2-39}
 & MLP & with & 51.19 & 50.05 & 48.00 & 38.34 & 30.08 & 17.24 & 50.92 & 49.07 & 47.31 & 38.09 & 29.17 & 17.29 & 48.83 & 47.38 & 45.56 & 36.15 & 30.07 & 18.58 & 44.89 & 43.24 & 41.20 & 32.92 & 24.76 & 17.47 & 42.87 & 41.90 & 39.75 & 32.47 & 26.55 & 18.63 & 48.48 & 47.08 & 44.76 & 37.61 & 29.30 & 20.56 \\
\cline{2-39}
 & \multirow[c]{2}{*}{RandomForest} & with & 54.79 & 52.91 & 51.61 & 41.66 & 33.86 & 19.23 & 55.18 & 53.02 & 51.44 & 40.69 & 33.62 & 20.06 & 53.87 & 51.82 & 49.53 & 40.27 & 33.48 & 21.44 & 50.85 & 48.98 & 46.88 & 38.01 & 28.90 & 20.07 & 50.85 & 48.30 & 45.97 & 38.33 & 30.33 & 19.72 & 53.78 & 51.56 & 49.84 & 42.43 & 34.47 & 20.96 \\
 &  & without & 56.06 & 53.66 & 50.92 & 41.57 & 32.56 & 21.50 & 55.68 & 54.34 & 51.71 & 43.96 & 35.53 & 21.96 & 55.88 & 53.80 & 51.42 & 44.15 & 37.11 & 23.81 & 53.71 & 52.77 & 51.04 & 45.27 & 34.95 & 24.30 & 52.98 & 51.57 & 50.21 & 44.08 & 35.23 & 23.12 & 55.63 & 54.60 & 53.38 & 45.07 & 38.30 & \textbf{25.91} \\
\cline{2-39}
 & SVM & with & 57.10 & 54.63 & 52.12 & 41.24 & 32.08 & 18.35 & 57.37 & 54.96 & 52.28 & 41.02 & 31.46 & 18.71 & 56.82 & 54.02 & 51.59 & 41.36 & 32.40 & 19.80 & 55.32 & 53.24 & 50.62 & 39.64 & 30.37 & 19.02 & 52.66 & 50.71 & 48.75 & 38.46 & 29.73 & 20.19 & 48.26 & 47.03 & 44.79 & 36.95 & 29.75 & 16.61 \\
\cline{2-39}
 & TABTransformer & with & 54.07 & 51.80 & 49.87 & 40.22 & 31.33 & 18.50 & 53.09 & 51.60 & 49.93 & 40.00 & 33.10 & 20.86 & 52.12 & 50.29 & 48.27 & 37.22 & 28.19 & 20.53 & 49.64 & 47.69 & 46.17 & 36.11 & 27.57 & 18.15 & 46.16 & 43.63 & 41.69 & 35.98 & 27.23 & 16.94 & 49.23 & 47.71 & 45.91 & 38.42 & 31.07 & 21.18 \\
\cline{2-39}
 & TabNet & with & 45.17 & 44.02 & 42.48 & 35.59 & 27.02 & 15.10 & 45.09 & 43.80 & 41.23 & 33.52 & 26.21 & 14.82 & 45.96 & 44.79 & 43.77 & 34.69 & 26.64 & 17.38 & 41.29 & 39.85 & 38.59 & 31.54 & 25.03 & 16.09 & 37.14 & 36.78 & 35.41 & 30.88 & 23.44 & 13.36 & 24.38 & 23.16 & 22.22 & 20.05 & 14.00 & 11.16 \\
\cline{2-39}
 & \multirow[c]{2}{*}{XGBoost} & with & 57.46 & 55.76 & 53.61 & 43.94 & 33.32 & 20.51 & 57.35 & 55.65 & 53.78 & 43.54 & 35.00 & 21.61 & 57.11 & 55.04 & 52.97 & 41.62 & 34.66 & 21.35 & 55.94 & 53.75 & 52.03 & 42.75 & 32.09 & 22.77 & 52.75 & 50.56 & 49.15 & 41.18 & 33.33 & 21.90 & 52.79 & 51.41 & 49.40 & 41.52 & 32.67 & 21.92 \\
 &  & without & 58.14 & 55.48 & 52.37 & 42.73 & 33.17 & 21.74 & 58.07 & 56.55 & 53.45 & 44.36 & 34.08 & 22.29 & 57.90 & 56.14 & 53.53 & 44.57 & 36.10 & 20.53 & 56.53 & 54.51 & 52.67 & 45.97 & 35.88 & 23.75 & 57.44 & \textbf{56.15} & 53.04 & 45.25 & 35.98 & 23.65 & 52.90 & 51.54 & 49.49 & 41.91 & 33.93 & 24.68 \\
\cline{1-39} \cline{2-39}
\multirow[c]{5}{*}{\textbf{Intermediate}} & FTTransformer & with & 0.65 & 0.59 & 0.52 & 0.39 & 0.29 & -0.15 & 0.26 & 0.31 & 0.27 & 0.37 & 0.20 & 0.15 & 4.67 & 4.48 & 4.30 & 3.72 & 2.98 & 1.42 & 0.00 & 0.00 & 0.00 & 0.00 & 0.00 & 0.00 & 0.00 & 0.00 & 0.00 & 0.00 & 0.00 & 0.00 & 0.00 & 0.00 & 0.00 & 0.00 & 0.00 & 0.00 \\
\cline{2-39}
 & MLP & with & 49.77 & 47.85 & 45.59 & 36.87 & 29.72 & 19.77 & 50.15 & 48.32 & 46.01 & 37.72 & 30.80 & 19.33 & 49.03 & 46.85 & 44.70 & 36.56 & 28.91 & 19.02 & 48.41 & 47.19 & 45.14 & 37.79 & 30.11 & 20.60 & 48.80 & 46.83 & 44.66 & 36.41 & 29.34 & 20.19 & 17.37 & 16.91 & 16.28 & 12.75 & 9.74 & 6.39 \\
\cline{2-39}
 & \modelName & without & \textbf{59.07} & \textbf{56.87} & 54.36 & \textbf{46.24} & \textbf{36.77} & 22.36 & \textbf{59.59} & \textbf{57.80} & \textbf{56.00} & 46.62 & 37.02 & 22.95 & \textbf{59.75} & \textbf{58.00} & \textbf{55.49} & 46.56 & 37.68 & 25.88 & \textbf{57.64} & \textbf{55.91} & \textbf{54.40} & \textbf{46.91} & 37.39 & 25.20 & \textbf{57.57} & 55.78 & \textbf{53.42} & \textbf{46.66} & 37.59 & 24.65 & \textbf{58.06} & \textbf{56.55} & \textbf{54.28} & \textbf{46.88} & \textbf{38.66} & 25.10 \\
\cline{2-39}
 & TABTransformer & with & 50.26 & 47.79 & 45.50 & 36.52 & 27.94 & 16.28 & 47.98 & 46.59 & 44.57 & 35.27 & 26.92 & 17.47 & 49.71 & 48.04 & 45.27 & 36.84 & 28.00 & 18.46 & 47.64 & 45.57 & 43.57 & 37.22 & 28.89 & 19.30 & 46.87 & 45.03 & 43.53 & 35.70 & 28.67 & 17.87 & 41.20 & 40.39 & 39.30 & 32.19 & 27.08 & 18.60 \\
\cline{2-39}
 & TabNet & with & 1.95 & 1.96 & 1.55 & 2.03 & 1.68 & 1.82 & 2.04 & 2.06 & 3.04 & 1.86 & 1.08 & 0.53 & 2.08 & 1.92 & 2.11 & 1.31 & 1.57 & -0.42 & 2.03 & 1.24 & 0.73 & 1.67 & 0.77 & 0.49 & 1.48 & 2.57 & 2.65 & 1.91 & 0.31 & -0.58 & 1.32 & 1.32 & 1.67 & 1.29 & 0.11 & -0.52 \\
\cline{1-39} \cline{2-39}
\multirow[c]{14}{*}{\textbf{Late}} & AdaBoost & with & 46.49 & 44.68 & 42.75 & 34.73 & 27.14 & 18.65 & 43.86 & 42.56 & 41.65 & 34.83 & 28.54 & 20.28 & 44.82 & 43.60 & 42.01 & 35.41 & 29.46 & 19.50 & 43.16 & 41.75 & 40.45 & 34.40 & 28.48 & 18.23 & 42.87 & 42.05 & 40.97 & 33.95 & 28.11 & 18.78 & 41.69 & 40.41 & 38.67 & 31.77 & 26.49 & 17.40 \\
\cline{2-39}
 & \multirow[c]{2}{*}{DecisionTree} & with & 30.13 & 30.30 & 28.81 & 25.00 & 18.51 & 12.43 & 27.96 & 28.31 & 27.87 & 24.68 & 22.64 & 20.06 & 29.76 & 29.22 & 27.65 & 24.42 & 23.17 & 21.65 & 28.12 & 27.74 & 26.87 & 23.08 & 19.08 & 17.15 & 29.35 & 29.37 & 28.32 & 24.91 & 23.62 & 21.69 & 27.68 & 27.02 & 26.15 & 23.30 & 20.97 & 17.80 \\
 &  & without & 38.39 & 37.38 & 35.48 & 29.32 & 22.65 & 11.68 & 40.89 & 40.48 & 40.32 & 36.93 & 32.36 & 27.28 & 40.57 & 39.73 & 38.60 & 33.98 & 30.04 & 24.47 & 39.11 & 38.12 & 38.12 & 35.12 & 31.34 & 25.27 & 37.20 & 36.69 & 35.59 & 32.32 & 28.67 & 24.90 & 36.02 & 35.56 & 35.40 & 31.61 & 28.04 & 22.58 \\
\cline{2-39}
 & FTTransformer & with & 47.62 & 45.40 & 43.05 & 34.46 & 27.28 & 14.05 & 47.50 & 45.63 & 43.60 & 34.70 & 26.93 & 15.08 & 47.70 & 45.75 & 43.73 & 34.90 & 27.78 & 16.39 & 46.71 & 45.37 & 43.33 & 35.85 & 28.51 & 17.52 & 47.12 & 45.78 & 43.67 & 36.13 & 28.95 & 18.60 & 22.96 & 22.64 & 21.89 & 17.71 & 13.92 & 8.78 \\
\cline{2-39}
 & \multirow[c]{2}{*}{HistGradientBoost} & with & 35.13 & 33.65 & 31.33 & 24.44 & 18.23 & 7.13 & 34.42 & 33.62 & 32.14 & 26.64 & 22.75 & 18.06 & 33.66 & 32.30 & 31.44 & 26.47 & 23.84 & 20.72 & 34.12 & 32.89 & 31.69 & 27.82 & 23.00 & 18.72 & 33.22 & 32.31 & 30.01 & 26.46 & 21.10 & 19.42 & 33.41 & 32.47 & 30.58 & 26.17 & 21.96 & 18.97 \\
 &  & without & 38.64 & 36.57 & 35.02 & 27.50 & 20.40 & 6.46 & 37.95 & 36.98 & 36.00 & 30.40 & 26.26 & 19.50 & 38.06 & 37.22 & 36.56 & 31.23 & 27.33 & 21.58 & 36.90 & 35.70 & 34.40 & 29.11 & 25.14 & 21.49 & 37.85 & 37.24 & 35.09 & 29.11 & 24.06 & 21.29 & 34.14 & 33.20 & 31.29 & 27.04 & 22.60 & 19.77 \\
\cline{2-39}
 & MLP & with & 43.24 & 41.57 & 39.71 & 32.88 & 24.55 & 11.78 & 43.35 & 41.85 & 40.09 & 33.48 & 26.84 & 17.30 & 43.36 & 42.11 & 40.04 & 34.20 & 28.14 & 18.10 & 42.70 & 41.32 & 39.57 & 34.43 & 27.69 & 18.63 & 43.67 & 42.32 & 41.01 & 34.81 & 27.94 & 19.78 & 45.80 & 44.39 & 42.55 & 34.98 & 27.72 & 19.19 \\
\cline{2-39}
 & \multirow[c]{2}{*}{RandomForest} & with & 43.16 & 41.24 & 38.55 & 30.62 & 22.72 & 10.44 & 43.12 & 42.67 & 41.97 & 36.97 & 32.15 & 25.73 & 42.16 & 41.56 & 40.63 & 38.22 & 32.74 & 26.42 & 41.97 & 41.63 & 40.91 & 36.37 & 30.38 & 24.98 & 42.12 & 41.48 & 40.15 & 36.54 & 29.79 & 22.22 & 39.48 & 38.84 & 37.46 & 32.71 & 27.88 & 21.42 \\
 &  & without & 51.15 & 49.50 & 47.27 & 40.30 & 30.21 & 15.15 & 54.03 & 53.07 & 51.68 & \textbf{46.93} & \textbf{41.32} & \textbf{31.64} & 52.99 & 52.42 & 51.18 & \textbf{47.62} & \textbf{42.38} & \textbf{32.16} & 52.00 & 51.24 & 50.26 & 46.82 & \textbf{41.70} & \textbf{31.15} & 51.21 & 50.59 & 49.49 & 45.74 & \textbf{39.99} & \textbf{30.35} & 46.93 & 45.92 & 44.85 & 40.80 & 34.37 & 24.86 \\
\cline{2-39}
 & SVM & with & 46.09 & 44.05 & 42.06 & 33.39 & 25.67 & 12.25 & 45.51 & 43.56 & 41.69 & 33.58 & 26.30 & 14.02 & 45.17 & 43.00 & 41.00 & 33.55 & 26.89 & 15.66 & 40.12 & 39.01 & 37.29 & 30.67 & 23.94 & 15.29 & 32.38 & 31.43 & 30.79 & 26.87 & 22.91 & 15.65 & 15.92 & 15.16 & 14.61 & 12.23 & 10.34 & 6.54 \\
\cline{2-39}
 & TABTransformer & with & 41.30 & 39.59 & 37.70 & 30.24 & 22.42 & 8.15 & 41.31 & 40.11 & 38.27 & 31.44 & 24.84 & 13.25 & 42.70 & 41.13 & 39.17 & 32.48 & 26.21 & 15.21 & 40.21 & 39.38 & 38.16 & 32.91 & 26.69 & 17.27 & 40.52 & 39.62 & 38.30 & 32.22 & 26.45 & 17.94 & 35.85 & 35.10 & 34.30 & 29.31 & 24.88 & 17.84 \\
\cline{2-39}
 & TabNet & with & 40.23 & 38.92 & 36.91 & 30.29 & 24.87 & 13.91 & 41.15 & 40.03 & 38.39 & 31.90 & 25.73 & 16.02 & 39.00 & 37.95 & 36.11 & 30.50 & 24.65 & 16.15 & 36.24 & 35.42 & 34.14 & 29.00 & 23.65 & 16.74 & 32.59 & 31.27 & 30.37 & 25.07 & 20.40 & 13.96 & 14.21 & 14.13 & 13.25 & 12.11 & 9.04 & 6.14 \\
\cline{2-39}
 & \multirow[c]{2}{*}{XGBoost} & with & 44.84 & 43.09 & 41.45 & 32.96 & 25.63 & 13.87 & 46.16 & 44.67 & 42.82 & 35.00 & 27.17 & 15.52 & 46.07 & 44.50 & 43.18 & 35.50 & 27.99 & 18.07 & 46.38 & 45.19 & 43.20 & 35.82 & 28.75 & 18.78 & 46.24 & 45.09 & 42.83 & 34.99 & 27.18 & 18.43 & 44.52 & 42.70 & 40.80 & 34.12 & 27.52 & 18.15 \\
 &  & without & 46.82 & 44.75 & 42.62 & 36.67 & 29.29 & 14.74 & 46.42 & 44.85 & 43.20 & 36.81 & 29.43 & 19.28 & 46.84 & 45.66 & 44.37 & 37.53 & 30.87 & 20.87 & 46.08 & 44.54 & 42.96 & 36.45 & 29.96 & 19.22 & 46.77 & 45.47 & 43.32 & 37.49 & 28.62 & 20.15 & 44.90 & 43.13 & 40.90 & 34.22 & 27.74 & 18.86 \\
\bottomrule
\end{tabular}
}
\captionof{table}{Average AUC and MCC performance of the experiments across the respective tasks in the ``missing modalities'' setting. 
To facilitate the analysis we highlighted the best performance in each column in bold.}
\label{tab:MOD_res}
\end{minipage}
\end{adjustbox}
\end{figure}
\vfill

\newpage

\vfill
\begin{figure}[!htbp]
\centering
\resizebox{.8\linewidth}{!}{
\begin{tabular}{lll|cccc|cccc|cccc|cccc}
\toprule
 \textbf{AUC} &  &  & \multicolumn{4}{c|}{\textbf{Train missing percentage:} $\Omega$} & \multicolumn{4}{c|}{\textbf{Train missing percentage:} 30\%} & \multicolumn{4}{c|}{\textbf{Train missing percentage:} 50\%} & \multicolumn{4}{c}{\textbf{Train missing percentage:} 75\%} \\
 &  &  & \multicolumn{4}{c|}{\textbf{Test missing percentage:}} & \multicolumn{4}{c|}{\textbf{Test missing percentage:}} & \multicolumn{4}{c|}{\textbf{Test missing percentage:}} & \multicolumn{4}{c}{\textbf{Test missing percentage:}} \\
\textbf{Fusion Strategy} & \textbf{Model} & \textbf{Imputer} & $\Omega$ & 30\% & 50\% & 75\% & $\Omega$ & 30\% & 50\% & 75\% & $\Omega$ & 30\% & 50\% & 75\% & $\Omega$ & 30\% & 50\% & 75\% \\
\midrule
\multirow[c]{14}{*}{\textbf{Early}} & AdaBoost & with & 83.89 & 77.53 & 82.18 & 78.04 & 76.65 & 76.10 & 73.36 & 68.80 & 84.44 & 78.21 & 81.80 & 78.00 & 81.66 & 72.82 & 80.07 & 76.00 \\
\cline{2-19}
 & \multirow[c]{2}{*}{DecisionTree} & with & 76.40 & 65.40 & 74.30 & 67.49 & 65.10 & 65.63 & 60.00 & 58.39 & 74.89 & 62.44 & 73.13 & 67.49 & 68.93 & 58.40 & 68.47 & 65.15 \\
 &  & without & 78.33 & 64.65 & 75.27 & 65.65 & 66.07 & 66.74 & 64.49 & 55.37 & 76.77 & 62.93 & 75.63 & 67.68 & 71.45 & 59.06 & 70.62 & 69.75 \\
\cline{2-19}
 & FTTransformer & with & 87.10 & 79.70 & 85.50 & 80.29 & 79.19 & 78.76 & 74.88 & 69.07 & 86.34 & 77.20 & 84.49 & 79.37 & 83.52 & 71.44 & 80.79 & 76.96 \\
\cline{2-19}
 & \multirow[c]{2}{*}{HistGradientBoost} & with & 88.05 & 78.35 & 86.40 & 80.06 & 79.34 & 78.84 & 75.44 & 69.79 & 87.90 & 77.65 & 85.73 & 79.81 & 85.73 & 74.23 & 83.66 & 78.90 \\
 &  & without & 89.99 & 80.59 & 87.27 & 78.82 & 81.67 & 80.54 & 75.29 & 67.26 & 89.93 & 80.70 & 88.07 & 82.05 & \textbf{89.10} & 78.54 & \textbf{87.56} & 82.62 \\
\cline{2-19}
 & MLP & with & 85.08 & 75.44 & 83.14 & 77.86 & 72.99 & 73.90 & 71.12 & 65.51 & 83.94 & 72.07 & 82.60 & 77.55 & 82.80 & 72.24 & 80.88 & 76.93 \\
\cline{2-19}
 & \multirow[c]{2}{*}{RandomForest} & with & 88.75 & 80.64 & 86.95 & 81.84 & 80.97 & 80.64 & 77.47 & 70.50 & 88.13 & 78.84 & 86.35 & 81.36 & 85.85 & 75.24 & 84.33 & 78.91 \\
 &  & without & 90.78 & \textbf{82.27} & \textbf{89.22} & \textbf{83.63} & \textbf{82.96} & \textbf{82.35} & \textbf{78.85} & \textbf{72.10} & 90.17 & \textbf{81.20} & \textbf{89.04} & \textbf{84.85} & 88.09 & \textbf{78.87} & 87.12 & 83.94 \\
\cline{2-19}
 & SVM & with & 86.43 & 75.79 & 84.62 & 79.75 & 76.49 & 75.93 & 72.47 & 65.13 & 86.66 & 74.99 & 84.51 & 79.54 & 85.57 & 74.34 & 83.10 & 78.37 \\
\cline{2-19}
 & TABTransformer & with & 84.87 & 73.84 & 83.03 & 77.86 & 72.35 & 72.23 & 69.82 & 64.22 & 84.24 & 73.18 & 82.53 & 78.17 & 82.43 & 71.14 & 80.54 & 75.68 \\
\cline{2-19}
 & TabNet & with & 85.11 & 75.86 & 84.11 & 78.72 & 76.03 & 75.67 & 74.13 & 68.04 & 84.93 & 75.74 & 82.95 & 77.78 & 80.52 & 68.61 & 79.22 & 73.15 \\
\cline{2-19}
 & \multirow[c]{2}{*}{XGBoost} & with & 88.00 & 78.30 & 86.12 & 79.19 & 78.65 & 78.14 & 73.92 & 68.52 & 87.15 & 76.14 & 85.15 & 78.99 & 84.42 & 71.01 & 82.86 & 78.53 \\
 &  & without & 90.07 & 79.44 & 86.55 & 77.30 & 80.16 & 79.14 & 75.04 & 68.34 & 89.22 & 78.17 & 87.83 & 82.61 & 86.89 & 74.42 & 86.76 & \textbf{84.16} \\
\cline{1-19} \cline{2-19}
\multirow[c]{5}{*}{\textbf{Intermediate}} & FTTransformer & with & 76.82 & 69.54 & 74.18 & 69.98 & 72.40 & 70.00 & 65.30 & 60.30 & 68.47 & 64.18 & 66.80 & 64.03 & 50.00 & 50.00 & 50.00 & 50.00 \\
\cline{2-19}
 & MLP & with & 86.61 & 78.94 & 84.96 & 81.16 & 80.63 & 79.60 & 75.42 & 69.67 & 82.58 & 67.02 & 81.44 & 78.70 & 75.61 & 50.00 & 75.61 & 73.68 \\
\cline{2-19}
 & \modelName & without & \textbf{90.80} & 80.60 & 87.81 & 82.75 & 82.31 & 81.29 & 74.02 & 65.10 & \textbf{90.24} & 80.06 & 87.80 & 82.13 & 88.21 & 74.30 & 85.58 & 81.07 \\
\cline{2-19}
 & TABTransformer & with & 85.32 & 75.91 & 83.66 & 79.54 & 73.95 & 72.88 & 69.73 & 63.43 & 83.90 & 73.93 & 82.22 & 78.21 & 81.46 & 67.70 & 79.43 & 76.11 \\
\cline{2-19}
 & TabNet & with & 75.58 & 69.67 & 73.57 & 70.89 & 69.88 & 68.47 & 64.16 & 59.89 & 73.29 & 63.37 & 72.14 & 70.10 & 67.17 & 62.60 & 67.72 & 63.89 \\
\cline{1-19} \cline{2-19}
\multirow[c]{14}{*}{\textbf{Late}} & AdaBoost & with & 83.94 & 78.32 & 81.73 & 76.69 & 77.61 & 77.39 & 73.46 & 67.79 & 83.50 & 76.31 & 81.90 & 78.41 & 81.86 & 73.34 & 80.38 & 77.72 \\
\cline{2-19}
 & \multirow[c]{2}{*}{DecisionTree} & with & 77.36 & 66.37 & 75.27 & 70.16 & 67.78 & 67.27 & 64.59 & 57.76 & 76.02 & 65.82 & 74.34 & 70.11 & 72.44 & 60.34 & 71.09 & 67.16 \\
 &  & without & 83.18 & 70.14 & 79.75 & 69.13 & 71.52 & 70.37 & 65.09 & 59.79 & 80.85 & 68.42 & 78.47 & 70.73 & 77.18 & 65.68 & 75.43 & 73.72 \\
\cline{2-19}
 & FTTransformer & with & 87.78 & 79.09 & 85.87 & 81.40 & 80.98 & 80.06 & 76.09 & 71.17 & 86.64 & 76.37 & 84.79 & 80.89 & 85.20 & 73.03 & 83.96 & 80.11 \\
\cline{2-19}
 & \multirow[c]{2}{*}{HistGradientBoost} & with & 83.29 & 73.04 & 80.87 & 75.24 & 73.52 & 73.26 & 69.43 & 64.36 & 82.30 & 71.83 & 80.34 & 75.68 & 80.74 & 69.54 & 78.87 & 75.05 \\
 &  & without & 84.66 & 73.00 & 80.95 & 73.02 & 74.27 & 73.65 & 66.82 & 62.13 & 84.25 & 73.65 & 81.71 & 74.83 & 82.48 & 71.05 & 79.93 & 75.12 \\
\cline{2-19}
 & MLP & with & 87.13 & 79.12 & 85.09 & 81.02 & 81.01 & 80.29 & 76.56 & 71.65 & 86.76 & 78.90 & 84.89 & 81.23 & 85.31 & 77.10 & 83.98 & 80.23 \\
\cline{2-19}
 & \multirow[c]{2}{*}{RandomForest} & with & 84.92 & 71.79 & 82.80 & 78.52 & 73.02 & 71.86 & 68.06 & 63.66 & 84.12 & 70.63 & 82.52 & 78.76 & 81.98 & 65.25 & 80.92 & 78.38 \\
 &  & without & 88.70 & 77.01 & 86.52 & 80.82 & 79.62 & 78.78 & 74.39 & 69.19 & 88.79 & 79.40 & 87.22 & 82.78 & 86.87 & 76.65 & 85.64 & 83.03 \\
\cline{2-19}
 & SVM & with & 88.13 & 78.98 & 86.20 & 81.69 & 80.33 & 79.16 & 76.01 & 70.70 & 87.82 & 78.98 & 85.96 & 82.00 & 86.78 & 78.21 & 85.44 & 81.93 \\
\cline{2-19}
 & TABTransformer & with & 85.46 & 75.05 & 83.28 & 79.44 & 77.69 & 77.03 & 72.83 & 67.32 & 85.03 & 75.37 & 83.20 & 79.77 & 82.54 & 70.90 & 80.95 & 77.77 \\
\cline{2-19}
 & TabNet & with & 87.12 & 78.08 & 85.03 & 81.09 & 78.46 & 77.40 & 73.67 & 69.37 & 86.43 & 76.42 & 84.90 & 80.81 & 83.60 & 72.95 & 82.20 & 79.62 \\
\cline{2-19}
 & \multirow[c]{2}{*}{XGBoost} & with & 87.65 & 77.55 & 85.37 & 79.86 & 78.79 & 78.32 & 74.07 & 67.20 & 87.32 & 77.01 & 85.18 & 80.59 & 85.06 & 72.99 & 83.55 & 79.69 \\
 &  & without & 89.05 & 78.21 & 86.13 & 76.76 & 80.63 & 80.26 & 75.00 & 68.33 & 88.34 & 77.70 & 87.20 & 81.87 & 86.68 & 75.02 & 85.97 & 83.27 \\
\bottomrule
\end{tabular}

}

\vspace*{.5cm}

\resizebox{.8\linewidth}{!}{
\begin{tabular}{lll|cccc|cccc|cccc|cccc}
\toprule
 \textbf{MCC} &  &  & \multicolumn{4}{c|}{\textbf{Train missing percentage:} $\Omega$} & \multicolumn{4}{c|}{\textbf{Train missing percentage:} 30\%} & \multicolumn{4}{c|}{\textbf{Train missing percentage:} 50\%} & \multicolumn{4}{c}{\textbf{Train missing percentage:} 75\%} \\
 &  &  & \multicolumn{4}{c|}{\textbf{Test missing percentage:}} & \multicolumn{4}{c|}{\textbf{Test missing percentage:}} & \multicolumn{4}{c|}{\textbf{Test missing percentage:}} & \multicolumn{4}{c}{\textbf{Test missing percentage:}} \\
\textbf{Fusion Strategy} & \textbf{Model} & \textbf{Imputer} & $\Omega$ & 30\% & 50\% & 75\% & $\Omega$ & 30\% & 50\% & 75\% & $\Omega$ & 30\% & 50\% & 75\% & $\Omega$ & 30\% & 50\% & 75\% \\
\midrule
\multirow[c]{14}{*}{\textbf{Early}} & AdaBoost & with & 54.79 & 42.40 & 51.87 & \textbf{46.94} & 44.12 & 44.11 & 40.29 & 30.76 & 53.53 & 45.20 & 50.99 & 42.47 & 51.08 & 42.22 & 46.91 & 37.64 \\
\cline{2-19}
 & \multirow[c]{2}{*}{DecisionTree} & with & 47.80 & 38.64 & 47.23 & 40.77 & 38.40 & 36.56 & 36.00 & 32.32 & 43.48 & 32.36 & 43.23 & 37.08 & 33.69 & 24.19 & 33.26 & 28.50 \\
 &  & without & 50.50 & 41.29 & 43.22 & 28.99 & 41.48 & 39.92 & 34.16 & 20.07 & 45.78 & 37.01 & 44.76 & 37.41 & 35.60 & 26.52 & 36.33 & 33.04 \\
\cline{2-19}
 & FTTransformer & with & 54.17 & 38.05 & 50.58 & 43.36 & 37.30 & 36.66 & 31.88 & 28.88 & 52.11 & 32.33 & 50.00 & 42.52 & 47.66 & 30.32 & 46.48 & 38.45 \\
\cline{2-19}
 & \multirow[c]{2}{*}{HistGradientBoost} & with & 58.61 & \textbf{47.79} & 53.01 & 44.53 & 47.86 & 45.95 & 40.78 & \textbf{32.98} & 57.44 & 46.68 & 54.51 & 44.51 & 54.05 & 41.73 & 50.22 & 40.28 \\
 &  & without & 57.42 & 43.88 & 49.58 & 29.49 & 47.14 & 45.64 & 41.46 & 22.76 & 57.88 & 44.28 & 55.38 & 44.82 & 56.04 & \textbf{44.07} & \textbf{54.26} & 45.10 \\
\cline{2-19}
 & MLP & with & 51.21 & 34.08 & 48.72 & 42.34 & 36.02 & 35.18 & 32.03 & 26.29 & 48.59 & 28.66 & 47.22 & 40.75 & 44.42 & 25.81 & 44.60 & 36.96 \\
\cline{2-19}
 & \multirow[c]{2}{*}{RandomForest} & with & 54.79 & 38.09 & 51.59 & 44.54 & 39.81 & 39.01 & 38.88 & 32.20 & 53.59 & 38.16 & 50.69 & 42.98 & 44.33 & 25.86 & 44.49 & 37.26 \\
 &  & without & 56.06 & 38.42 & 50.28 & 33.00 & 42.12 & 40.37 & 31.33 & 20.58 & 52.59 & 34.93 & 51.20 & 41.21 & 46.08 & 25.57 & 45.29 & 42.13 \\
\cline{2-19}
 & SVM & with & 57.10 & 41.61 & 53.19 & 44.17 & 43.28 & 41.26 & 39.00 & 29.67 & 54.38 & 37.35 & 51.54 & 42.02 & 52.83 & 34.05 & 50.11 & 37.68 \\
\cline{2-19}
 & TABTransformer & with & 54.49 & 42.26 & 50.37 & 42.80 & 39.70 & 39.38 & 32.66 & 26.18 & 50.57 & 34.41 & 48.52 & 40.81 & 46.82 & 28.53 & 44.49 & 37.93 \\
\cline{2-19}
 & TabNet & with & 46.36 & 30.30 & 43.46 & 37.48 & 31.89 & 30.90 & 28.81 & 21.53 & 45.93 & 29.30 & 43.93 & 35.34 & 36.05 & 23.08 & 36.13 & 27.17 \\
\cline{2-19}
 & \multirow[c]{2}{*}{XGBoost} & with & 57.46 & 44.66 & 53.51 & 45.20 & 46.22 & \textbf{46.28} & \textbf{41.62} & 32.04 & 56.68 & 45.40 & 52.90 & \textbf{45.54} & 53.47 & 40.92 & 51.37 & 40.75 \\
 &  & without & 58.14 & 46.00 & 47.47 & 27.78 & \textbf{48.00} & 45.58 & 40.06 & 18.42 & 56.21 & 41.52 & 54.83 & 42.64 & 53.35 & 40.86 & 53.05 & \textbf{45.43} \\
\hline
\multirow[c]{5}{*}{\textbf{Intermediate}} & FTTransformer & with & 1.51 & 3.78 & 0.95 & 0.54 & 1.06 & 1.06 & 0.34 & 0.36 & 0.00 & 0.00 & 0.00 & 0.00 & 0.00 & 0.00 & 0.00 & 0.00 \\
\cline{2-19}
 & MLP & with & 49.13 & 30.24 & 49.06 & 43.20 & 30.27 & 30.27 & 29.74 & 25.54 & 49.06 & 29.98 & 47.34 & 40.56 & 44.52 & 26.83 & 43.10 & 37.52 \\
\cline{2-19}
 & \modelName & without & \textbf{59.95} & 47.06 & \textbf{53.88} & 42.60 & 47.77 & 44.40 & 38.09 & 25.97 & \textbf{59.57} & \textbf{48.00} & \textbf{55.39} & 40.96 & \textbf{56.42} & 41.22 & 52.40 & 42.27 \\
\cline{2-19}
 & TABTransformer & with & 49.81 & 32.64 & 48.40 & 42.15 & 34.27 & 34.99 & 31.17 & 22.42 & 47.44 & 31.33 & 45.46 & 38.93 & 41.22 & 24.66 & 39.51 & 35.37 \\
\cline{2-19}
 & TabNet & with & 3.96 & 5.72 & 3.85 & 1.51 & 0.90 & 2.22 & 1.12 & 1.13 & 4.68 & 3.10 & 3.34 & 1.10 & 1.18 & 2.51 & 1.65 & 3.08 \\
\cline{1-19} \cline{2-19}
\multirow[c]{14}{*}{\textbf{Late}} & AdaBoost & with & 46.42 & 30.37 & 45.47 & 41.85 & 31.14 & 30.32 & 28.16 & 25.78 & 45.78 & 32.32 & 44.54 & 38.01 & 37.39 & 29.56 & 37.58 & 32.85 \\
\cline{2-19}
 & \multirow[c]{2}{*}{DecisionTree} & with & 30.16 & 27.84 & 27.18 & 21.15 & 28.41 & 27.06 & 22.40 & 16.46 & 27.99 & 25.08 & 25.78 & 20.66 & 18.89 & 17.24 & 19.61 & 16.24 \\
 &  & without & 38.39 & 35.36 & 31.03 & 21.85 & 34.26 & 33.29 & 25.21 & 16.56 & 33.85 & 32.32 & 31.72 & 21.83 & 23.22 & 20.32 & 23.73 & 23.37 \\
\cline{2-19}
 & FTTransformer & with & 47.53 & 29.40 & 45.98 & 37.91 & 29.13 & 29.13 & 27.19 & 19.10 & 47.32 & 28.52 & 44.35 & 35.38 & 43.20 & 27.42 & 41.10 & 33.14 \\
\cline{2-19}
 & \multirow[c]{2}{*}{HistGradientBoost} & with & 35.13 & 22.53 & 32.12 & 26.85 & 25.70 & 26.32 & 22.32 & 13.38 & 35.17 & 27.19 & 30.50 & 22.35 & 30.55 & 19.08 & 26.95 & 22.00 \\
 &  & without & 38.64 & 27.10 & 29.57 & 16.51 & 30.44 & 24.62 & 18.54 & 9.88 & 37.66 & 28.90 & 35.78 & 25.60 & 32.54 & 25.40 & 32.38 & 25.31 \\
\cline{2-19}
 & MLP & with & 43.73 & 27.41 & 42.31 & 36.72 & 26.24 & 26.24 & 24.08 & 21.16 & 43.05 & 25.42 & 41.41 & 33.73 & 40.70 & 24.28 & 38.85 & 32.08 \\
\cline{2-19}
 & \multirow[c]{2}{*}{RandomForest} & with & 43.04 & 32.86 & 41.00 & 33.13 & 33.82 & 33.07 & 29.92 & 19.00 & 41.99 & 30.20 & 39.18 & 30.88 & 34.38 & 22.08 & 30.81 & 24.97 \\
 &  & without & 51.15 & 38.32 & 44.81 & 23.88 & 41.10 & 39.10 & 28.09 & 9.27 & 48.43 & 33.86 & 44.40 & 30.03 & 40.88 & 26.74 & 39.77 & 36.04 \\
\cline{2-19}
 & SVM & with & 46.09 & 27.90 & 44.58 & 35.87 & 27.80 & 27.90 & 26.28 & 17.45 & 45.15 & 27.70 & 43.31 & 33.60 & 42.64 & 26.28 & 39.93 & 31.09 \\
\cline{2-19}
 & TABTransformer & with & 42.25 & 25.60 & 40.67 & 35.87 & 25.24 & 25.40 & 24.04 & 19.02 & 40.01 & 22.42 & 37.77 & 31.07 & 37.42 & 23.32 & 35.89 & 30.19 \\
\cline{2-19}
 & TabNet & with & 39.52 & 23.30 & 39.00 & 32.32 & 23.50 & 24.38 & 22.82 & 18.08 & 37.61 & 19.68 & 36.40 & 29.02 & 31.43 & 18.09 & 29.39 & 24.57 \\
\cline{2-19}
 & \multirow[c]{2}{*}{XGBoost} & with & 44.84 & 27.55 & 42.74 & 37.98 & 30.74 & 29.78 & 24.76 & 23.25 & 46.09 & 29.40 & 43.86 & 33.62 & 42.39 & 28.90 & 40.64 & 33.60 \\
 &  & without & 46.82 & 30.09 & 37.94 & 20.43 & 30.09 & 27.89 & 22.04 & 10.45 & 47.55 & 28.77 & 45.78 & 33.15 & 44.14 & 29.60 & 43.39 & 37.15 \\
\bottomrule
\end{tabular}
}
\captionof{table}{Average AUC and MCC performance of the experiments across the respective tasks in the ``all missing'' setting. 
To facilitate the analysis we highlighted the best performance in each column in bold.}
\label{tab:ALL_res}
\end{figure}
\vfill

\end{document}